\documentclass[10pt,twocolumn,letterpaper]{article}

\usepackage[pagenumbers]{cvpr} 

\usepackage{booktabs}
\usepackage{multirow}
\usepackage{multicol}
\usepackage{tcolorbox}

\usepackage[table]{xcolor} 
\definecolor{tablegreen}{HTML}{E6F7E6} 
\newcommand{\modelopt}{DOCR-Inspector\xspace}
\newcommand{\datasetopt}{DOCRcase-200K\xspace}
\newcommand{\benchopt}{DOCRcaseBench\xspace}

\newcommand{\redtext}[1]{\textcolor{red}{#1}}

\usepackage{tabularx}
\usepackage{soul, color, xcolor}
\usepackage{adjustbox}
\usepackage{colortbl}

\usepackage{float}
\usepackage{fancyvrb}
\usepackage{listings}
\usepackage{longtable}
\makeatletter

\newcommand{\Rmnum}[1]{\expandafter\@slowromancap\romannumeral #1@}
\makeatother

\definecolor{cvprblue}{rgb}{0.21,0.49,0.74}
\usepackage[pagebackref,breaklinks,colorlinks,allcolors=cvprblue]{hyperref}

\def\paperID{9736} 
\def\confName{CVPR}
\def\confYear{2026}

\title{DOCR-Inspector: Fine-Grained and Automated Evaluation of Document Parsing with VLM}

\author{
Qintong Zhang$^{*1,2}$\quad
Junyuan Zhang$^{*3}$\quad
Zhifei Ren$^{2}$\quad
Linke Ouyang$^{2}$\quad
Zichen Wen$^{4}$\quad\\
Junbo Niu$^{2}$\quad
Yuan Qu$^{2}$\quad
Bin Wang$^{2}$\quad
Ka-Ho Chow$^{3}$\quad
Conghui He$^{2\ddag}$\quad
Wentao Zhang$^{1\ddag}$\quad\\
$^1$Peking University\quad
$^2$Shanghai AI Laboratory\quad\\
$^3$The University of HongKong\quad
$^4$Shanghai Jiaotong University\quad
}

\begin{document}
\maketitle

\begin{abstract}
Document parsing aims to transform unstructured PDF images into semi-structured data, facilitating the digitization and utilization of information in diverse domains.
While vision language models (VLMs) have significantly advanced this task, achieving reliable, high-quality parsing in real-world scenarios remains challenging.
Common practice often selects the top-performing model on standard benchmarks.
However, these benchmarks may carry dataset-specific biases, leading to inconsistent model rankings and limited correlation with real-world performance.
Moreover, benchmark metrics typically provide only overall scores, which can obscure distinct error patterns in output.
This raises a key challenge: how can we reliably and comprehensively assess document parsing quality in the wild?
We address this problem with DOCR-Inspector, which formalizes document parsing assessment as fine-grained error detection and analysis.
Leveraging VLM-as-a-Judge, DOCR-Inspector analyzes a document image and its parsed output, identifies all errors, assigns them to one of 28 predefined types, and produces a comprehensive quality assessment.
To enable this capability, we construct DOCRcase-200K for training and propose the Chain-of-Checklist reasoning paradigm to enable the hierarchical structure of parsing quality assessment.
For empirical validation, we introduce DOCRcaseBench, a set of 882 real-world document parsing cases with manual annotations.
On this benchmark, DOCR-Inspector-7B outperforms commercial models like Gemini 2.5 Pro, as well as leading open-source models. 
Further experiments demonstrate that its quality assessments provide valuable guidance for parsing results refinement, making DOCR-Inspector both a practical evaluator and a driver for advancing document parsing systems at scale.
Model and code are released at: \url{https://github.com/ZZZZZQT/DOCR-Inspector}.
\end{abstract}


\begin{figure}[h]
  \centering
   \includegraphics[width=1.0\linewidth]{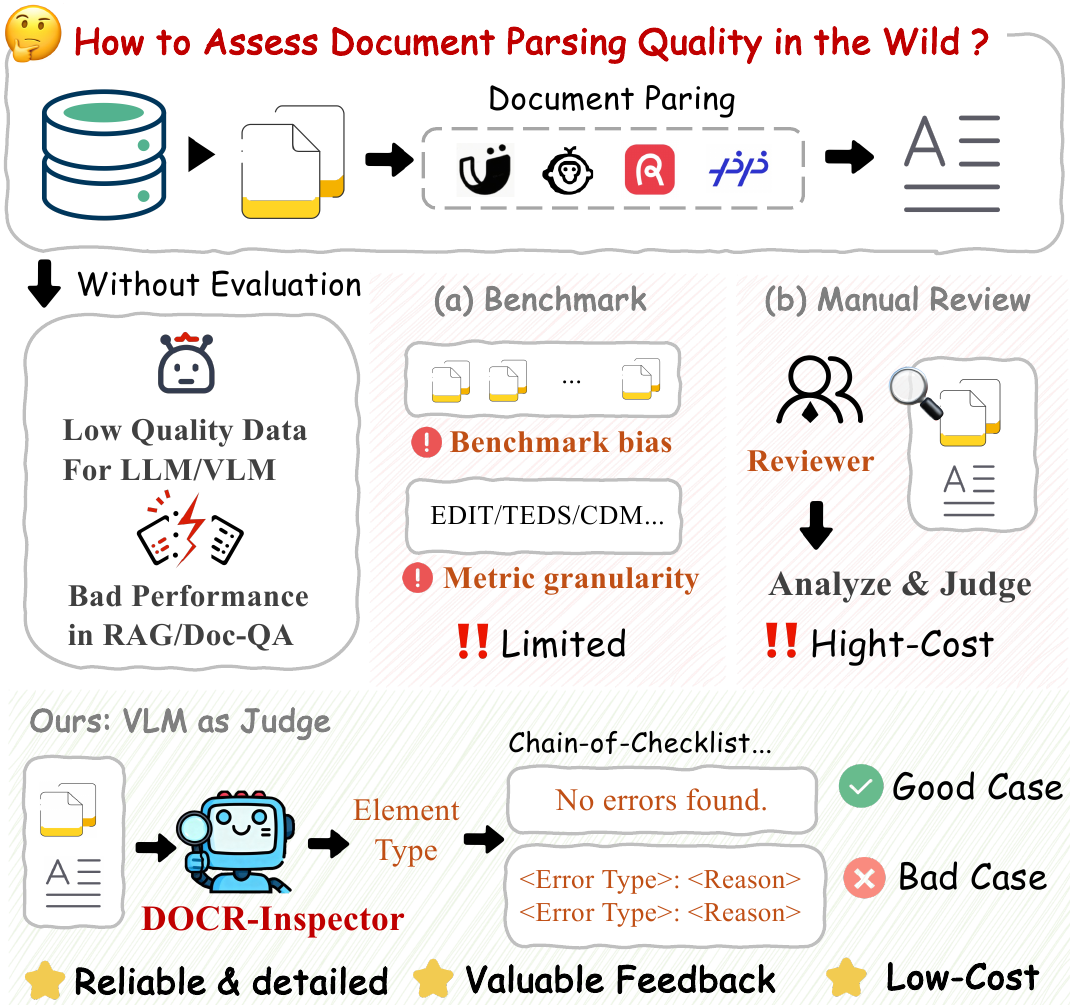}
    \vspace{-6mm}
   \caption{\textbf{Overview of Document Parsing Quality Assessment Methods.} 
   \modelopt uses Vision-Language Model(VLM) for reliable and detailed error detection, offering quality assessment as valuable feedback that guides parsing result refinement.}
   \label{fig:intro}
   \vspace{-6mm}
\end{figure}

\section{Introduction}
\label{sec:intro}

Document parsing aims to transform unstructured document images into semi-structured representations~\cite{zhang2024document}, supporting a wide range of downstream applications such as constructing knowledge-rich corpora for large language model (LLM) pretraining~\cite{liu2025points, xia2024docgenome,wang2024unimernet,hwang2021spatial, an2025concept} and building domain-specific knowledge bases for information retrieval~\cite{zhang2025ocr,dong2025doc}.
Recent advances in pipeline-based document parsing tools~\cite{wang2024mineru} and specialized large vision language models (VLMs)~\cite{wei2024general, niu2025mineru2, cui2025paddleocr}, have substantially advanced parsing accuracy.
Nevertheless, parsing real-world documents remains challenging due to their visual and structural diversity.
Pipeline systems often accumulate errors when handling out-of-domain inputs (e.g., newspapers or slides), leading to missing or misinterpreted content~\cite{ouyang2025omnidocbench}. 
Meanwhile, VLM-based approaches are prone to hallucinations, especially when processing low-quality or degraded documents~\cite{he2025seeing}.
As these errors propagate to downstream~\cite{zhang2025ocr}, there is an urgent need for reliable quality evaluation methods to bridge the gap between imperfect parsing outputs and the increasing demand for accurate, application-ready document data.

Currently, efforts to bridge the gap have largely focused on developing reliable benchmarks~\cite{poznanski2025olmocr, ouyang2025omnidocbench} that estimate performance using overall scores, with the goal of identifying the most suitable model for high-quality parsing results.
However, this evaluation paradigm faces two critical limitations in document parsing:
\textbf{(1) Benchmark bias.} 
Model rankings often vary between benchmarks due to differences in data distribution and evaluation method, causing inconsistent conclusions across test suites. 
Moreover, overall scores provide only a single performance estimate over a dataset, overlooking variations in parsing difficulty across individual documents. 
This disconnect can cause benchmark results to diverge from actual performance in deployment, as we expect to achieve high-quality parsing for each document.
\textbf{(2) Limited metric granularity.} Overall scores provide a convenient summary but fail to reveal systematic error patterns~\cite{li2025score,chen2025logics}.
For example, missing paragraphs and minor character-level noise can yield similar edit-distance scores, yet have profoundly different consequences for downstream tasks.
Manual inspection can reveal such patterns, but it is infeasible for large-scale pipelines.  
These limitations make it difficult to assess parsing quality in real-world settings, where ground truth data is unavailable and document-level performance is critical.

\begin{figure*}[!htbp]
  \centering
   \includegraphics[width=0.98\linewidth]{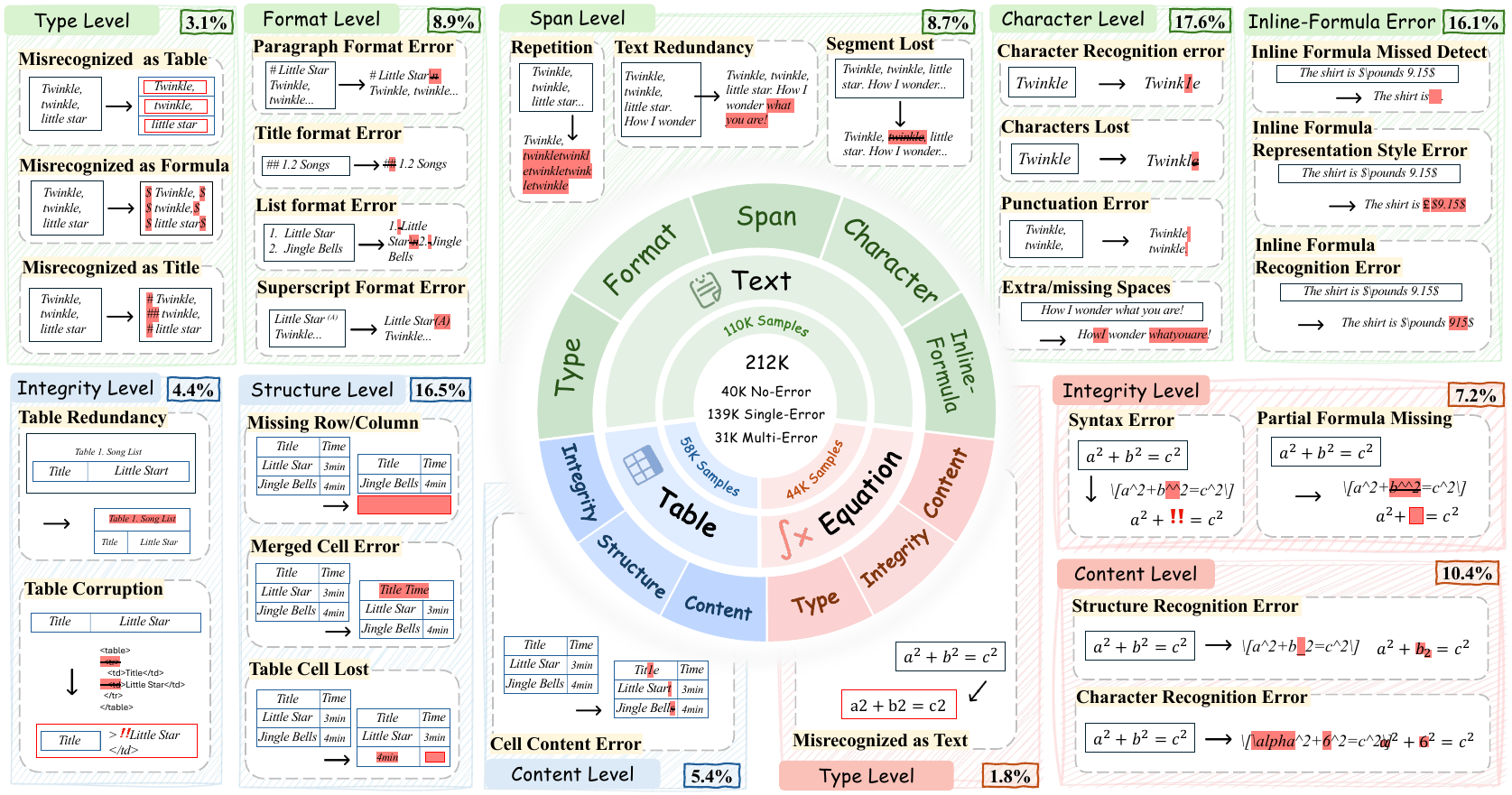}
   \vspace{-3mm}
   \caption{Overview of Error Type Definition and Distribution in \datasetopt. Our \datasetopt comprises 212k element images and parsing instances with error annotations, including 110k text, 58k table, and 44k equation elements. It defines 28 fine-grained error types organized into 11 main error levels. Criteria for these error types can be found in Appendix ~\cref{appendix:error_type_definition}.}
   \label{fig:datasets}
   \vspace{-5mm}
\end{figure*}

To address these challenges, we propose \modelopt, a VLM-based evaluation framework that provides a comprehensive and detailed quality assessment for document parsing results without ground truth.
\modelopt identifies all parsing errors and categorizes them into fine-grained error types, producing a comprehensive and detailed quality assessment.
As illustrated in~\cref{fig:datasets}, our error taxonomy includes 28 error types organized across three element categories and 11 main error levels, derived from extensive empirical analysis of real-world parsing outputs. 
We further introduce \datasetopt, a large-scale dataset containing diverse document elements, parsing results, and fine-grained error annotations, and instantiate \modelopt-7B as the evaluation model.

A central challenge in developing \modelopt\ is to capture multi-level errors while maintaining a low false-positive rate.
To this end, we introduce the Chain-of-Checklist (CoCL) reasoning paradigm, a structured process that guides the model through hierarchical verification from coarse to fine granularity, producing reliable and interpretable judgments.
We further adopt a two-stage training strategy that combines supervised fine-tuning (SFT) with group relative policy optimization (GRPO)~\cite{shao2024deepseekmath} under an asymmetric reward design, effectively balancing missed detections and misclassifications.

Our main contributions are as follows:
\begin{enumerate}
\item We develop a structured taxonomy for document parsing errors and release \datasetopt, a synthetic dataset annotated with fine-grained error types and reasoning traces.
\item We propose \modelopt and introduce \benchopt, a benchmark comprising 882 real-world document elements with human annotations. \modelopt-7B achieves superior performance in parsing quality evaluation, surpassing leading proprietary models and open-source VLMs, including Gemini 2.5 Pro Thinking and Qwen3-VL-235B-A22B-Thinking.
\item We demonstrate that quality assessment provided by \modelopt can guide the refinement of parsing outputs across multiple public benchmarks effectively, highlighting its practical value for improving real-world document parsing systems.
\end{enumerate}

\section{Related Work}
\label{sec:relatedwork}

\subsection{Document Parsing and Evaluation}
Document parsing has evolved rapidly, driven by advances in both pipeline-based systems and VLMs.
Early studies such as MinerU~\cite{wang2024mineru} and PaddleOCR~\cite{cui2025paddleocr} adopt modular pipelines combining layout detection, table recognition and formula recognition.
Correspondingly, most benchmarks focus on evaluating individual modules~\cite{zhao2024doclayout,wang2024unimernet,zhong2020image}.
While this design offers flexibility, it suffers from error accumulation across modules, hindering robustness on complex real-world documents.
This renders the evaluation results of a single module ineffective in the results of full-page document analysis, as downstream modules, such as table recognition, are affected by the accuracy of upstream modules like layout detection.
Recent research has shifted toward unified VLM-based approaches~\cite{bai2025qwen2,wei2024general,niu2025mineru2,feng2025dolphin,li2025monkeyocr,niu2025native}, which leverage multimodal understanding to perform end-to-end document parsing.
Existing benchmarks provide broad performance comparisons but suffer from biases, both across different benchmarks and between benchmarks and real-world scenarios~\cite{chen2025ocean,poznanski2025olmocr,ouyang2025omnidocbench,fu2024ocrbench} . Furthermore, their overall metrics make it difficult to localize and categorize specific error types, hindering the acquisition of actionable insights into parsing quality.
To address these challenges, we introduce \modelopt, a label-free framework for reliable and fine-grained evaluation of document parsing results in the wild.

\subsection{VLM-as-a-Judge}

The rapid progress of large language models (LLMs) has significantly improved open-ended tasks such as writing, while also revealing the limitations of traditional reference-based metrics like F1 and BLEU in evaluating generated responses~\cite{liu2023g, zheng2023judging, kim2023prometheus,lin2025perceive}.
These metrics focus on surface-level token overlap and fail to capture semantic quality or reasoning coherence.
As a result, an increasing number of studies are beginning to explore using large language models as evaluators, thereby reducing reliance on human assessment~\cite{lee2024prometheus, he2025audiomarathon}.
Extending this paradigm, VLMs are increasingly used as evaluators in multimodal domains such as synthetic image detection, visual question answering, and harmful content detection~\cite{kang2025legion,ku2023viescore, zhang2023gpt,yin2024woodpecker,lu2025vlm, niu2025ovo}.
However, these applications typically address coarse-grained or binary decisions focused on quality control or safety verification rather than detailed analytical evaluation.
To evaluate and obtain better document parsing results, 
DianJin-OCR-R1~\cite{chen2025dianjin} refines its initial outputs by referencing parsing results from other models.
While this improves accuracy to some extent, document parsing inherently requires comprehensive, hierarchical evaluation—spanning diverse sub-tasks (e.g., formula and table recognition) and multiple error granularity, from character-level misreads to structural inconsistencies.
To address this, we formulate document quality assessment as an error detection task and introduce Chain-of-Checklist (CoCL), a structured reasoning paradigm tailored to document hierarchies. 
With CoCL, our \modelopt generates a comprehensive and detailed quality assessment, offering deep insight into document parsing performance.

\section{\datasetopt}
\label{sec:dataset}
In this section, we formally define the task of document parsing quality evaluation (\cref{sec:definition}) and detail the construction of \datasetopt (\cref{sec:dataset_constrcution}), a large-scale dataset designed for fine-grained error detection and analysis. 
\datasetopt contains 212K element-level parsing cases spanning 28 error types across text, table and equation elements; each error is paired with detailed reasoning annotations. 
In \cref{checklist} we present the Chain-of-Checklist (CoCL) reasoning paradigm (\cref{fig:checklist}) used in our framework.

\subsection{Problem Definition}
\label{sec:definition}
We formulate document parsing quality evaluation as an error type detection task. Formally, given a document image $\text{Doc}$ containing a set of element regions $\mathcal{C} = \{crop_1, \dots, crop_n\}$ and a parsing model $M$, we obtain a set of parsing outputs $\text{pred}_{doc}$. The operation of $M$ is defined as:

\vspace{-4mm}
\begin{equation}
\begin{split}
\label{eq:parsing_output}
\text{pred}_{doc} &= M\left(\text{Doc}, \mathcal{C}\right) \\ 
&= \{ \text{pred}_{crop_1}, \dots, \text{pred}_{crop_n} \}
\end{split}
\end{equation}

The primary objective is to evaluate these outputs. For each region-result pair $X_i = (crop_i, \text{pred}_{crop_i})$, the task is to identify the error type $E_{crop_i} \in \mathcal{E}$ present in $\text{pred}_{crop_i}$, where $\mathcal{E}$ is the predefined error taxonomy.
To establish an objective and fine-grained error taxonomy $\mathcal{E}$, we first categorized document elements into three fundamental types: text, tables, and equations. Based on manual analysis of parsing outputs across diverse real-world documents, we developed a two-level taxonomy comprising 11 major error levels and 28 specific fine-grained error types, as detailed in \cref{fig:datasets}.
Complete definitions and illustrative examples for each error type are provided in appendix. 

\begin{figure}[t]
  \centering
   \includegraphics[width=1.0\linewidth]{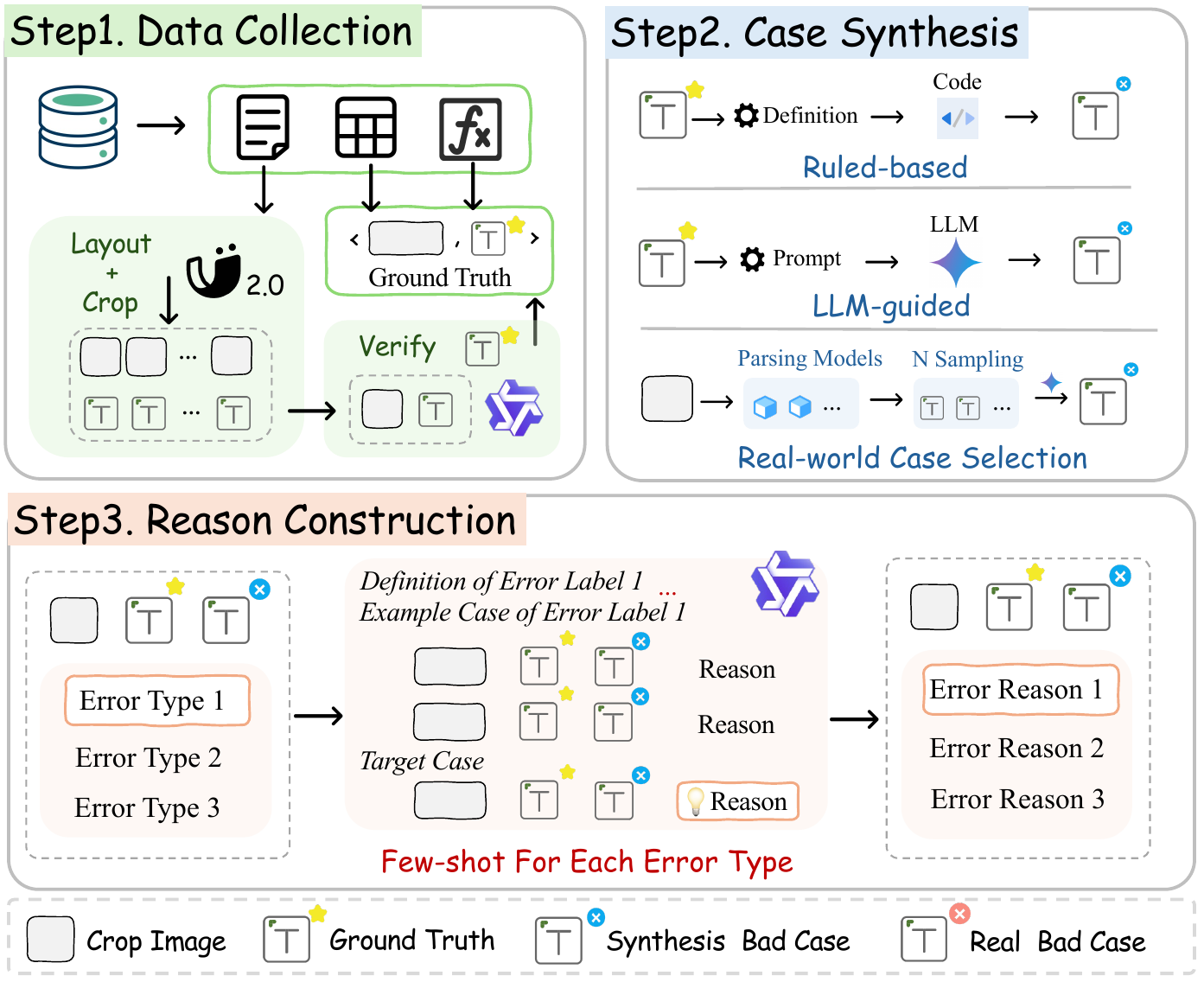}
   \vspace{-5mm}
   \caption{Overview of the \datasetopt construction pipline.}
   \label{fig:pipeline}
   \vspace{-5mm}
\end{figure}
\subsection{\datasetopt Construction}
\label{sec:dataset_constrcution}
As illustrated in \cref{fig:pipeline}, \datasetopt is constructed through a multi-stage pipeline designed to reflect real-world parsing error distributions while maintaining high-quality annotations.

\noindent\textbf{Data Collection.}
To capture diverse layouts, languages, and domains, we sample full-page documents from two primary corpora: olmOCR-mix-1025~\cite{olmOCR-mix-0225} (5,000 randomly selected samples) and the CDLA corpus~\cite{li2021cdla}. Because equations and tables are underrepresented in full pages, we augment the collection with 10K equation instances from UniMER-1M~\cite{wang2024unimernet} and 10K table instances from internal repositories to mitigate structural imbalance.
To obtain precise element-level ground truth, we extract element-level crops using MinerU2.0-vlm~\cite{MinerU20} for layout and initial parsing, followed by a refinement pass using Qwen2.5-VL-72B-Instruct~\cite{bai2025qwen2} to correct each element crop.
For UniMER-1M and high-quality internal table annotations, we directly reuse the provided ground truth.

\noindent\textbf{Case Synthesis.}Existing synthesis methods, such as REVISE~\cite{shim2025revise}, primarily rely on rule-based perturbations, which fail to capture semantic and structurally complex errors commonly observed in real-world document parsing. To ensure balanced coverage across all 28 error types, we design three complementary synthesis strategies based on ground truth. Detailed procedures are described in the appendix.

\begin{enumerate}
    \item \textbf{Rule-Based Synthesis}: We generate simple and pattern-driven errors by applying deterministic perturbations to the ground truth.
    
    \item \textbf{LLM-guided Synthesis}: To simulate model-induced errors such as hallucinations, semantic inconsistencies, and structural distortions that are difficult to reproduce with fixed rules, we employ targeted prompts to guide Gemini 2.5 Flash in generating realistic erroneous outputs from the ground truth.
    
    \item \textbf{Real-world Case Selection}: For complex or severe error patterns that rule-based or LLM-guided methods fail to reproduce consistently, we curate representative failure cases from real-world parsing outputs.
\end{enumerate}

To reflect the cases with multiple errors of real-world parsing systems, we strategically combine error synthesis methods to construct cases containing 2–4 non-interfering error types.
The final dataset includes 212,424 instances—110,040 text, 57,510 table, and 44,874 equations samples. The proportions of error-free \textit{Good Case}, \textit{Bad Case} with single error and \textit{Bad Case} with multiple error are 18.83\%, 65.17\%, and 16.00\%, respectively, as illustrated in~\cref{fig:datasets}. Comprehensive statistics for \datasetopt are presented in appendix.

\subsection{Chain-of-Checklist}
\label{checklist}
\begin{figure}[t]
  \centering
   \includegraphics[width=1.0\linewidth]{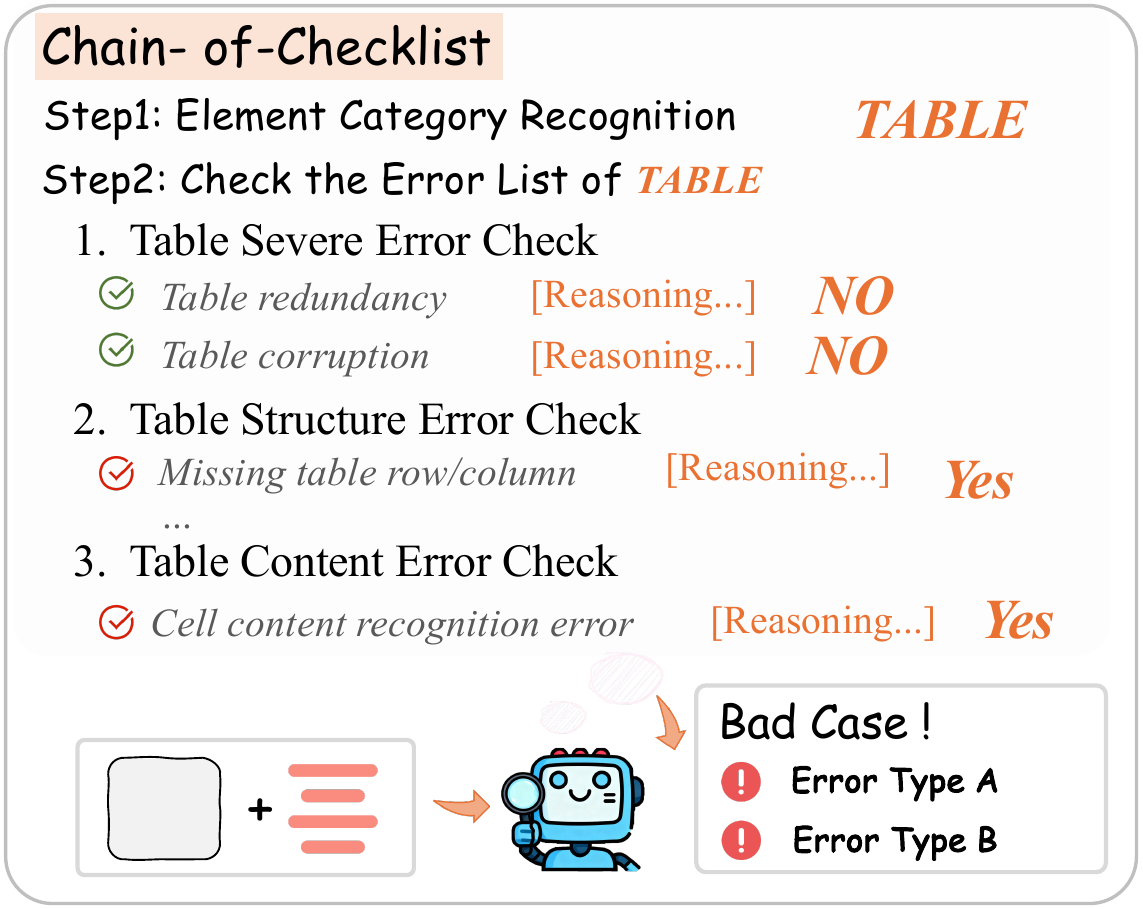}
   \vspace{-5mm}
   \caption{The Chain-of-Checklist (CoCL) reasoning template. 
   The complete templates for all elements are provided in the appendix.
   }
   \label{fig:checklist}
   \vspace{-5mm}
\end{figure}

As shown in~\cite{su2025learning,lu2025vlm}, reasoning improves a model’s ability to identify hallucinations, reducing the number of required sampling iterations.
However, in error type detection for document parsing, we found that despite using CoT reasoning mode, the model frequently struggled to identify and distinguish error types accurately, particularly in cases containing not only one error.

To overcome this limitation, we introduce the Chain-of-Checklist (CoCL) reasoning paradigm, which systematically guides the model through a comprehensive image-text alignment and analysis process for error detection. As illustrated in \cref{fig:checklist}, CoCL employs a structured checklist tailored for each document element type, enumerating potential errors specific to that category.
Samples with detected errors are labeled as \textit{Bad Cases} with corresponding error tags, while error-free samples are designated as \textit{Good Cases}. This structured design enforces comprehensive inspection and substantially improves the model’s ability to detect subtle or compound errors.

Constructing reliable reasoning chains for CoT remains challenging~\cite{jiang2025mme,yu2025rethinking}, often yielding inconsistent or noisy supervision.
To build high-quality CoCL reasoning data, we generate reasoning chains for each error type independently and subsequently merge them into unified CoCL representations. Specifically, a few-shot prompting strategy is used to guide Qwen2.5-VL-72B-Instruct to focus on one error type per case and produce its reasoning trace. 
The resulting chains are aggregated into the CoCL format, forming a coherent and interpretable reasoning annotation.
This synthesis process produces independent reasoning annotations that can be flexibly composed into structured reasoning chains, resulting in high-quality CoCL reasoning data.
\section{\modelopt-7B}
\label{sec:method}
This section describes how we train \modelopt-7B using our proposed \datasetopt, starting from Qwen2.5-VL-7B-Instruct as the base model.
The overall training pipeline consists of two stages. The first stage adopts SFT to gradually adapt Qwen2.5-VL-7B-Instruct from a general-purpose VLM in a specialized juder for document parsing quality. The second stage applies GRPO with asymmetric rewards to further enhance the evaluation accuracy of \modelopt.
Additional details are provided in the appendix.

\noindent\textbf{SFT with CoCL.} 
In the first stage, the model is trained on the full \datasetopt\ to acquire advanced quality assessment capabilities and learn the Chain-of-Checklist (CoCL) reasoning paradigm. During this phase, the language model, visual encoder, and alignment module of Qwen2.5-VL-7B-Instruct are all fine-tuned, allowing the model to jointly optimize multimodal representations and better adapt to element-level feature extraction. 

\noindent\textbf{GRPO with Asymmetric Reward.} 
Following SFT, we observed that fine-tuning across all samples causes the model to become overly sensitive—often predicting spurious errors and misclassifying \textit{Good Cases} as \textit{Bad Cases}.
Therefore, the model is further optimized using Generative Reward Policy Optimization (GRPO) in the second stage. To strategically balance the goals of improving recall for error types in \textit{Bad Cases} and minimizing the misidentification of \textit{Good Cases}, we define an asymmetric reward function as $R$:

\vspace{-4mm}
\begin{equation}
R =
\begin{cases}
\mathcal{S}_{\text{Format}} + \mathcal{S}_{\text{F1}} + \mathcal{S}_{\text{Recall}}, & \text{Badcase,} \\
\mathcal{S}_{\text{Format}} + \mathcal{S}_{\text{Precision}}, & \text{Goodcase.}
\end{cases}  
\end{equation}

This formulation explicitly differentiates between cases:
for \textit{Bad Cases}, $\mathcal{S}_{\text{F1}}$ and $\mathcal{S}_{\text{Recall}}$ encourage comprehensive error coverage and reduce missed detections;
for \textit{Good Cases}, $\mathcal{S}_{\text{Precision}}$ penalizes false positives, preventing over-sensitivity and overfitting.
The $\mathcal{S}_{\text{Format}}$ term ensures structural consistency and adherence to output formatting standards.
For GRPO training, we select approximately 3k high-quality samples from \datasetopt based on the model’s error detection performance after the first stage, ensuring a balanced mix of \textit{Good Case} and \textit{Bad Case} instances. 
This stage continues from the SFT checkpoint, and only the language model component of the model is fine-tuned, while the visual encoder 
and alignment module remains frozen.
\section{Experiments}
\label{sec:experiments}
\begin{table}[!t]
  \centering
  \caption{Distribution of each element cases in \benchopt} 
  \label{tab:bench}
  \small
  \begin{tabular}{l c c c} 
    \toprule 
    & \textbf{text} & \textbf{table} & \textbf{equation} \\
    \midrule 
    Good Case & 39 & 46& 62\\
    Bad Case with Single Error & 339 & 141&  81\\
    Bad Case with Multi Error & 70 &55&  49\\
    \textbf{Total} & 448& 242 & 192\\
    \bottomrule 
  \end{tabular}
  \vspace{-5mm}
\end{table}
\begin{table*}[t]
  \centering
  \caption{Results of document parsing quality assessment on \benchopt. 
  We report F1 for case quality judgment accuracy. For document parsing error type detection, we report F1, precision, and recall across text, table, and equation. 
  Bold indicates the best performance, and underline indicates the second-best performance.}
  \label{tab:model_performance}
  \footnotesize
  \renewcommand\arraystretch{0.95}
  \setlength{\tabcolsep}{4 pt}
  \begin{tabular}{c *{12}{c}}
    \toprule
   \multirow{2}{*}{\textbf{Model}} & \multicolumn{4}{c}{\textbf{Text}} & \multicolumn{4}{c}{\textbf{Table}} & \multicolumn{4}{c}{\textbf{Equation}} \\
   \cmidrule(lr){2-5} \cmidrule(lr){6-9} \cmidrule(lr){10-13} 
   & \multicolumn{1}{c|}{\textbf{Case}} & \multicolumn{3}{c|}{\textbf{Error Type}} & \multicolumn{1}{c|}{\textbf{Case}} & \multicolumn{3}{c|}{\textbf{Error Type}} & \multicolumn{1}{c|}{\textbf{Case}} & \multicolumn{3}{c}{\textbf{Error Type}} \\
   \cmidrule(lr){2-2} \cmidrule(lr){3-5} \cmidrule(lr){6-6} \cmidrule(lr){7-9} \cmidrule(lr){10-10} \cmidrule(lr){11-13}
   & \textbf{F1} & \textbf{Recall} & \textbf{F1} & \textbf{Precision} & \textbf{F1} & \textbf{Recall} & \textbf{F1} & \textbf{Precision} & \textbf{F1} & \textbf{Recall} & \textbf{F1} & \textbf{Precision} \\

    \midrule
    \multicolumn{13}{l}{\textit{Proprietary Non-Reasoning Models}} \\
    \midrule
       GPT-4o w/o CoT & 72.05 & 31.66 & 28.8 & 28.04 & 73.69 & 29.89 & 26.36 & 25.03 & 79.2 & 49.31 & 47.2 & 46.31 \\
       GPT-4o w/ CoT & 77.69 & 30.54 & 27.25 & 26.35 & 81.23 & 34.23 & 29.64 & 28.17 & 79.38 & 46.44 & 45.45 & 45.4 \\
       Gemini 2.5 Flash w/o CoT  & 84.89 & 43.29 & 29.88 & 25.43 & 82.21 & 41.94 & 25.97 & 21.29 & 80.46 & 53.73 & 48.17 & 45.96 \\
       Gemini 2.5 Flash w/ CoT & 84.75 & 42.24 & 29.74 & 25.69 & 81.16 & 42.36 & 24.1 & 19.25 & \underline{80.94} & 50.61 & 46.17 & 44.63 \\
    \midrule
    \multicolumn{13}{l}{\textit{Open-source Non-Reasoning Models}} \\
    \midrule
       Qwen2.5-VL-7B-Instruct w/o CoT & 46.15 & 12.28 & 11.98 & 11.83 & 48.8 & 19.42 & 19.42 & 19.42 & 55.8 & 32.81 & 32.81 & 32.81 \\
       Qwen2.5-VL-7B-Instruct w/ CoT & 38.17 & 12.05 & 11.64 & 11.50 & 43.48 & 21.56 & 21.72 & 22.11 & 68.1 & 32.29 & 32.12 & 32.03 \\
       Qwen2.5-VL-72B-Instruct w/o CoT & 82.68 & 28.49 & 24.74 & 23.43 & \underline{83.51} & 40.91 & 33.94 & \underline{31.03} & 78.51 & 39.93 & 37.19 & 35.76 \\
       Qwen2.5-VL-72B-Instruct w/ CoT & 74.55 & 30.97 & 26.23 & 24.56 & 76.82 & 40.70 & 31.77 & 28.43 & 79.14 & 44.53 & 41.23 & 39.79 \\
    \midrule
    \multicolumn{13}{l}{\textit{Reasoning Models}} \\
    \midrule
      Qwen3-VL-235B-A22B-Thinking & 83.9 & 42.02 & 31.19 & 27.46 & 83.13 & 39.12 & 28.57 & 25.49 & 78.56 & 40.8 & 38.45 & 37.76 \\
      Gemini 2.5 Pro Thinking  & \underline{88.46} & \underline{47.17} & \underline{32.9} & \underline{28.16} & 82.01 & \underline{43.60} & \underline{32.93} & 29.63 & 77.19 & 53.04 & \underline{48.58} & \underline{47.27} \\
    \midrule
    \multicolumn{10}{l}{\textit{Ours:}} \\
    \midrule
    \rowcolor{tablegreen} 
    DOCR-Inspector-7B & \textbf{96.43} & \textbf{81.06} & \textbf{80.21} & \textbf{81.03} & \textbf{86.41} & \textbf{63.09} & \textbf{62.11} & \textbf{62.95} & \textbf{85.42} & \textbf{74.39} & \textbf{73.81} & \textbf{74.48} \\
    \bottomrule
  \end{tabular}
  \vspace{-5mm}
\end{table*}

In the section, we first introduce \benchopt(\cref{sec:bench}), a benchmark of real-world document parsing samples with human annotations, and conduct comprehensive experiments to evaluate the effectiveness and practical utility of the proposed \modelopt framework and its resulting \modelopt-7B model.
We compare \modelopt-7B with leading proprietary and open-source VLMs to evaluate its performance in document parsing quality assessment on \benchopt (\cref{sec:performance})
Ablation studies (\cref{sec:Ablation}) further dissect the contribution of the CoCL reasoning paradigm and the asymmetric rewards, revealing how they enhance model performance. 
Finally, we analyze the correlation between assessment and metrics, and discuss the practical utility of \modelopt's evaluation for refining parsing results (\cref{sec:analyse}).
All prompts used in experiments are provided in the appendix.

\subsection{\benchopt}
\label{sec:bench}
We construct \benchopt, a high-quality benchmark specifically for evaluating document quality assessment. 
It comprises parsed outputs from several models, including MinerU2.0-pipeline~\cite{wang2024mineru}, PP-StructureV3~\cite{cui2025paddleocr}, GPT-4o~\cite{GPT-4o}, Qwen2.5-VL-7B-Instrcut~\cite{bai2025qwen2}, MonkeyOCR-1.2B-Pro~\cite{li2025monkeyocr}, and MinerU2.0-vlm~\cite{wang2024mineru}, which perform well but not the best in most of the benchmarks.
The construction process involves two main steps. First, we perform these models' inference on OmniDocBench-v1.0~\cite{ouyang2025omnidocbench} and apply its matching algorithm to align parsed outputs with element blocks and corresponding ground truth.
Second, we use the metrics to initially filter out \textit{Good Cases} and \textit{Bad Cases}.
All \textit{Bad Cases} are then manually reviewed by three trained annotators, who refine the error-type labels and augment the dataset to ensure a balanced distribution across element categories and error types. 
The overall composition of \benchopt\ is summarized in \cref{tab:bench}.

\subsection{Performance of \modelopt}
\label{sec:performance}
\noindent\textbf{Metrics.} On \benchopt, we evaluate model performance using multiple complementary metrics. 
Binary classification (\textit{Good Case} vs. \textit{Bad Case}) is determined by the presence of any predicted error, with the F1 used to assess overall quality judgment accuracy. 
For fine-grained error type detection, we report precision, recall, and F1.

\noindent\textbf{Models.} We considered three categories of models: (1) Proprietary non-reasoning models, including GPT-4o and Gemini 2.5 Flash~\cite{comanici2025gemini}. (2) Open-source non-reasoning models, such as Qwen2.5-VL-7B-Instruct and Qwen2.5-VL-72B-Instruct~\cite{bai2025qwen2}. (3)  Reasoning models, including Qwen3-VL-235B-A22B-Thinking~\cite{qwen3technicalreport} and Gemini 2.5 Pro Thinking~\cite{comanici2025gemini}. Each model is guided by a prompt that defines all error types to ensure consistent evaluation.

~\cref{tab:model_performance} presents a comprehensive comparison between \modelopt-7B and leading VLMs, and \modelopt-7B achieves the best performance across all document element types. 
In binary quality assessment (Case F1), \modelopt-7B surpasses the powerful reasoning model Gemini 2.5 Pro Thinking by significant margins of +7.97\%, +4.4\%, and +8.23\% on text, table, and formula tasks, respectively. 
A larger performance gap emerges in fine-grained error type detection. 
As shown in \cref{tab:model_performance}, neither non-reasoning models with CoT prompting nor advanced reasoning models achieve satisfactory recall, with most remaining below 50\%.
For example, on text, Gemini 2.5 Pro Thinking attains only 32.9\% F1.
This suggests that although general-purpose VLMs can handle coarse quality assessment, they struggle to identify subtle inconsistencies between document element images and parsed content, such as list formatting mistakes in text or merged cell misrecognitions in table.
Although reasoning capabilities enhance VLM performance, nearly all models remain inadequate for this highly specialized task.
In contrast, \modelopt-7B improves this metric remarkably to 80.21\% (+47.31\%). Similar substantial gains are observed in table (+29.18\% F1) and equation (+25.23\% F1) tasks. 
These results underscore the value of \datasetopt in training specialized evaluators and demonstrate the superiority of \modelopt in fine-grained, reliable document parsing quality assessment.

\subsection{Ablation studies}
\label{sec:Ablation}
We conduct ablation experiments on \benchopt to analyze the contribution of each component in \modelopt, using Qwen2.5-VL-7B-Instruct as the base model.
As shown in ~\cref{fig:checklist} and ~\cref{tab:Ablation}, we systematically evaluate the two-stage training strategy, focusing on the effectiveness of the CoCL reasoning paradigm and the asymmetric reward.

\begin{figure}[!t]
  \centering
   \includegraphics[width=1.0\linewidth]{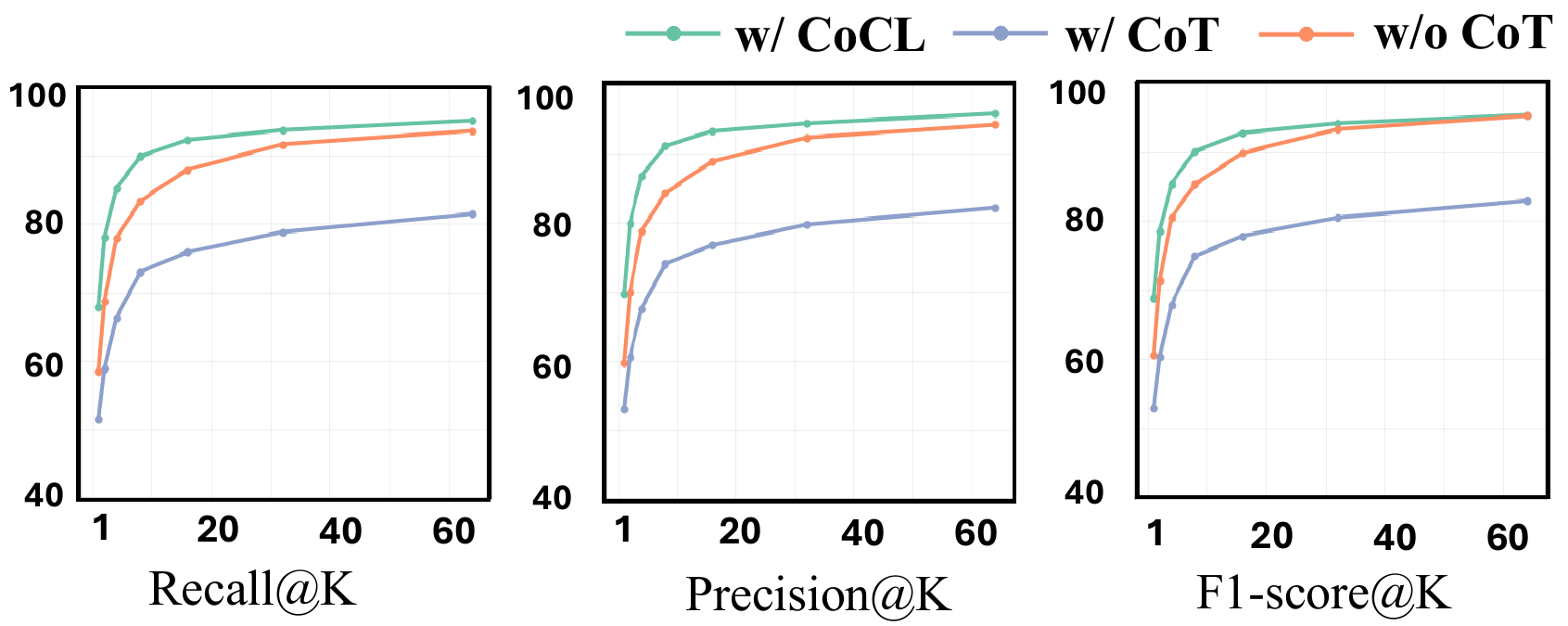}
   \vspace{-5mm}
   \caption{Pass@K of SFT stage model with different reasoning mode. We report Recall@K, Precision@K, and F1@K for K sampling times with a temperature setting of 1.0.
   }
   \vspace{-5mm}
   \label{fig:line_fig}
\end{figure}

\noindent\textbf{Effectiveness of CoCL}
To assess the impact of CoCL, we compare models trained on \datasetopt in the SFT stage using three reasoning modes: non-reasoning (w/o CoT), standard Chain-of-Thought (w/ CoT), and Chain-of-Checklist (w/ CoCL).
The multiple sampling results, shown in ~\cref{fig:line_fig}, reasoning significantly enhances performance, with CoCL providing the largest gain. 
Increasing the sampling number $K$ consistently preserves this advantage, suggesting that structured reasoning improves stability and reduces uncertainty during error detection.
Quantitatively, CoCL raises the Error Type F1 from 57.12\% (CoT) to 71.96\%, demonstrating a substantial improvement in fine-grained error.
The structured checklist process in CoCL significantly improves $pass@1$ by providing a more comprehensive and reliable reasoning pathway for error type detection

\noindent\textbf{Effectiveness of Asymmetric Reward.}
The CoCL reasoning paradigm established during the SFT stage provides a strong foundation for precise, fine-grained error detection. Building on this, we further optimize the model using GRPO with two reward schemes: a conventional F1-based reward and the proposed asymmetric reward. \modelopt-7B trained with the asymmetric reward achieves the highest overall performance, improving Case F1 from 90.22\% to 91.28\% and maintaining balanced Error Type Recall (74.68\%) and Precision (74.64\%). These results demonstrate that the asymmetric reward effectively mitigates over-sensitivity to $Good Case$ while preserving error detection capability, thereby enhancing both the accuracy and robustness of document parsing quality assessment.

\subsection{Analysis Experiments}
\label{sec:analyse}
To validate the effectiveness and practical value of the quality assessments provided by \modelopt, we designed two sets of experiments.
First, in ~\cref{sec:relationship}, we systematically analyze error patterns in the samples, examining the relationships between error frequency, error types, and performance metrics to validate the rationality of our proposed error taxonomy. 
Then, in \cref{sec:refinement}, we assess whether the quality assessment of \modelopt can serve as guidance to refine document parsing results, demonstrating their practical value beyond evaluation.

\noindent\textbf{Experimental Setup.}
To ensure diversity across element types, we curated text blocks (15–250 characters) from OmniDocBench-v1.5 (16,240 samples)~\cite{ouyang2025omnidocbench}, included the table recognition subset from CC-OCR~\cite{yang2025cc}, and randomly selected 360 samples from the SPE subset of UniMERTest~\cite{wang2024unimernet}.
We selected three leading document parsing models based on their performance on OmniDocBench-v1.5 as of October 1, 2025: 
dots.ocr~\cite{dotsocr}, 
MonkeyOCR-3B Pro~\cite{li2025monkeyocr}, 
and MinerU2.5~\cite{niu2025mineru2}.
For text recognition, we calculate edit distance with the text format. For table recognition, we use TEDS~\cite{zhong2020image} and S-TEDS, and use CDM~\cite{wang2025image} and edit distance for formula recognition.

\begin{table}[!t]
    \centering
    \caption{Ablation of training components on \benchopt. 
    We ablate the reasoning mode in the SFT stage and the reward function in the GRPO stage. 
    }
    \vspace{-3mm}
    \resizebox{\linewidth}{!}{
    \begin{tabular}{clcccc}
    \toprule
    \multirow{2}{*}{\textbf{Stage}} & \multirow{2}{*}{\textbf{Method}} & \multirow{2}{*}{\textbf{Case F1}} & \multicolumn{3}{c}{\textbf{Error Type}} \\
    \cmidrule(lr){4-6} 
    & & & \textbf{Recall} & \textbf{F1} & \textbf{Precision} \\
    \midrule
    \multicolumn{2}{c}{Qwen2.5-VL-7B-Instruct} & 47.80 & 18.71 & 18.56 & 18.48 \\
    \midrule
    \multirow{3}{*}{SFT} & w/o CoT & 83.91 & 52.56 & 51.34 & 52.31 \\
    & w/ CoT & 87.44 & 58.94 & 57.12 & 58.21 \\
    & \cellcolor{tablegreen}w/ CoCL & \cellcolor{tablegreen} 89.11 & \cellcolor{tablegreen} 71.77 & \cellcolor{tablegreen} 71.96 & \cellcolor{tablegreen} 73.50 \\
    \midrule
    \multirow{2}{*}{GRPO} & F1 Reward & 90.22 & 74.32 & 73.62 & 74.39 \\
    & \cellcolor{tablegreen}Asymmetric Reward& \cellcolor{tablegreen} 91.28 & \cellcolor{tablegreen} 74.68 & \cellcolor{tablegreen} 73.85 & \cellcolor{tablegreen} 74.64 \\
    \bottomrule
    \end{tabular}}
    \label{tab:Ablation}
    \vspace{-4mm}
\end{table}

\begin{figure*}[h]
  \centering
   \includegraphics[width=1.0\linewidth]{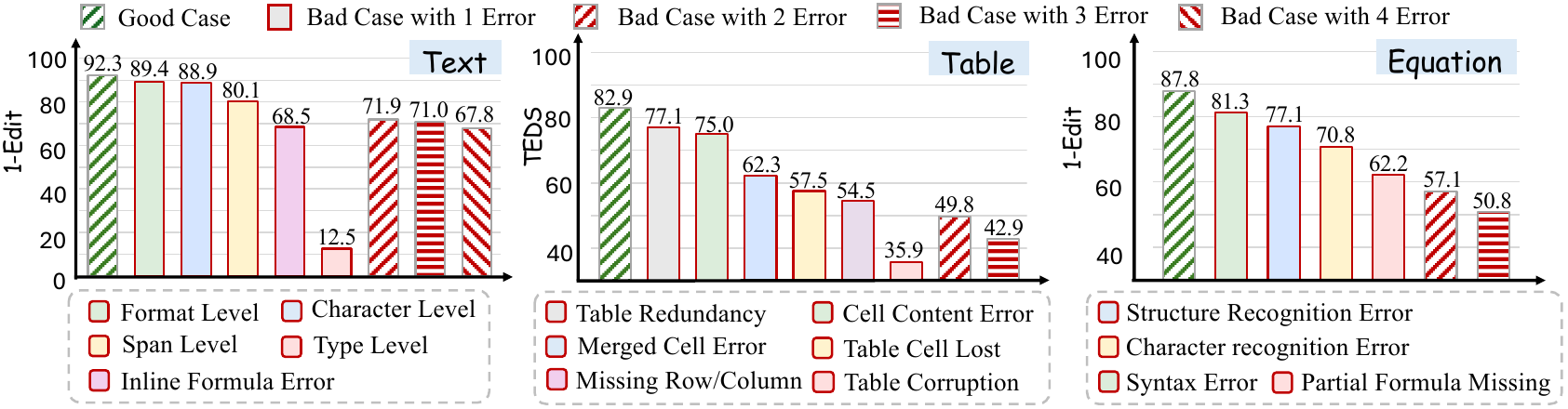}
   \vspace{-5mm}
   
   \caption{Performance metrics for three document elements under different error conditions. The bars represent the performance of cases classified by DOCR-Inspector as \textit{Good Case} and \textit{Bad Case} with 1 to 4 error types, with a detailed focus on the metrics of Bad Case samples containing different individual error types. 
   }
   \label{fig:bar_fig}
   \vspace{-5mm}
\end{figure*}

\subsubsection{Alignment with Objective Metrics}
\label{sec:relationship}
While ~\cref{sec:performance} establishes \modelopt's has better alignment with human judgment than others, a reliable VLM-as-a-Judge must also correlate with objective metrics~\cite{lee2024prometheus}.
We generated 10 sample sets using MinerU2.5 at a temperature of 1.2, ensuring broad coverage of diverse error types.
As illustrated in \cref{fig:bar_fig}, \modelopt-7B reliably reflects parsing quality—cases labeled as \textit{Good Case} consistently achieve the highest metric scores, while performance declines sharply as the number of error types increases.
For cases with a single error, the impact on metrics depends on the error’s type and level.
\textit{Bad Case} with minor format or character-related errors—such as Format Level and Character Level errors in text, or Cell Content errors in tables—exhibit only slight metric degradation. In contrast, cases involving structural or other critical errors—like Table Corruption in tables or Partial Formula Missing in equations—lead to substantial performance deterioration.


\subsubsection{Refinement with Guidance}
\label{sec:refinement}
As an effective evaluator, it is supposed to serve as a reward model, providing informative feedback to guide system refinement~\cite{wei2025perception,hu2024visual,luo2025language}. 
To examine the practical value of \modelopt-7B’s assessments, we design a refinement experiment in which Qwen2.5-VL-72B-Instruct serves as the refiner. The refiner receives cropped element images and initial parsing results from dots.ocr and MonkeyOCR-3B-Pro as inputs, and refines them under three guidance configurations: (1)\textbf{W}ith\textbf{o}ut \textbf{G}uidance (w/o G): The refiner autonomously evaluates and corrects all parsing outputs. (2)\textbf{W}ith \textbf{B}inary \textbf{G}uidance (w/ BG): The refiner is provided with coarse quality labels (“Good” or “Bad”) and refines only the bad cases. (3)\textbf{W}ith \textbf{D}etailed \textbf{G}uidance (w/ DG): The refiner leverages full and fine-grained assessment from \modelopt-7B to refine bad cases.

\begin{table}[h]
    \centering
    \caption{Refinement performance under different settings.}
    \vspace{-3mm}
    \label{tab:improvement}
    \small
    \resizebox{\linewidth}{!}{
    \begin{tabular}{m{1.2cm} c c c c c }
        \toprule
        \multirow{2}{*}{\textbf{Method}} & \textbf{Text} & \multicolumn{2}{c}{\textbf{Table}} & \multicolumn{2}{c}{\textbf{Equation}} \\
        \cmidrule(lr){2-2} 
        \cmidrule(lr){3-4} 
        \cmidrule(lr){5-6} 
        & Edit$\downarrow$ & TEDS$\uparrow$ & STEDS$\uparrow$ & CDM$\uparrow$ & Edit$\downarrow$\\
        \midrule
        
        \multicolumn{6}{l}{\textit{dots.ocr}} \\
        \midrule
        Initial & 16.3 & 75.4 & 81.7 & 96.9 & 22.7 \\
        
        w/o G& 11.1(\redtext{-5.2}) & 78.9(\redtext{+3.5}) & 84.1(\redtext{+2.4}) & 97.4(\redtext{+0.5}) & 14.9(\redtext{-7.8}) \\
        
        w/ BG & 11.6(\redtext{-4.7}) & 79.2(\redtext{+3.8}) & 84.4(\redtext{+2.7}) & 97.6(\redtext{+0.7}) & 15.7(\redtext{-7.0}) \\

        \rowcolor{tablegreen}
        w/ DG & 10.5(\redtext{-5.8}) & 79.8(\redtext{+4.4}) & 85.4(\redtext{+3.7}) & 98.5(\redtext{+1.6}) & 13.0(\redtext{-9.7}) \\
        
        \midrule
        
        \multicolumn{6}{l}{\textit{MonkeyOCR-3B-Pro}} \\
        \midrule
        Initial & 9.8 & 73.0 & 79.1 & 97.3 & 11.3 \\
        
        w/o G & 10.5(+0.7) & 79.9(\redtext{+6.9}) & 85.8(\redtext{+6.7}) & 98.4(\redtext{+1.1}) & 11.5(+0.2) \\
        
        w/ BG & 9.9(+0.1) & 79.9(\redtext{+6.9}) & 85.5(\redtext{+6.4}) & 98.4(\redtext{+1.1}) & 11.6(+0.3) \\
        
        \rowcolor{tablegreen} 
        w/ DG & 8.9(\redtext{-0.9}) & 80.3(\redtext{+7.3}) & 85.6(\redtext{+6.5}) & 98.6(\redtext{+1.3}) & 10.5(\redtext{-0.8}) \\
        \bottomrule
    \end{tabular}}
    \vspace{-6mm}
\end{table}

\begin{figure}[!t]
  \centering
   \includegraphics[width=1.0\linewidth]{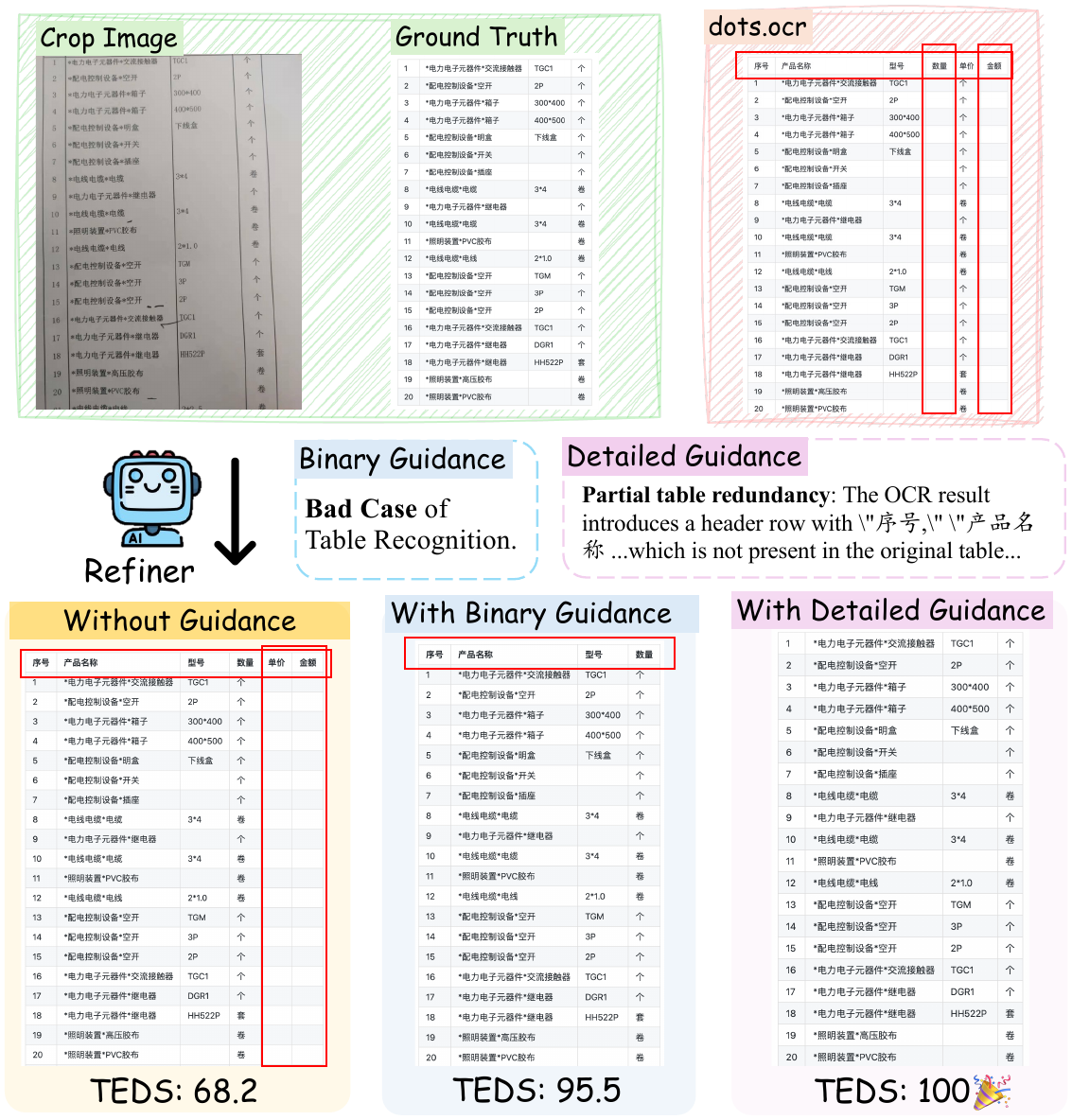}

   \caption{Case study of refinement.}
   \label{fig:case_study}
   \vspace{-6mm}
\end{figure}

As summarized in \Cref{tab:improvement}, Detailed Guidance yields consistently superior performance. This configuration produces the most substantial improvements across both base models. For instance, with Detailed Guidance, the text edit distance for MonkeyOCR-3B-Pro drops from 9.80 to 7.85, and the CDM score for dots.ocr on equation parsing improves by approximately 1.6, clearly surpassing the other settings.
We show a case in \cref{fig:case_study}, the initial output from dots.ocr contained redundant rows and columns. Without guidance, the refiner only adjusted column spacing without resolving the underlying structural issues. Under binary guidance, the refiner recognized the presence of table errors and removed extra blank columns, yet failed to address the redundant row. In contrast, detailed feedback precisely identified all error locations, enabling the refiner to fully reconstruct the correct table structure.
These results highlight the critical role of comprehensive and fine-grained quality assessment in model self-improvement. Detailed guidance not only steers refiners toward error-prone regions but also elevates the overall correction quality, demonstrating that \modelopt-7B can function as effective supervisory signals for iterative refinement.

\section{Conclusion}
\label{sec:conclusion}
In this paper, we propose \modelopt, a VLM-based framework for fine-grained and label-free quality assessment of document parsing results in real-world scenarios. 
To overcome the limitations of benchmark bias and limited metric granularity, we establish a structured error taxonomy and construct \datasetopt, a large-scale benchmark annotated with 28 detailed error types and corresponding reasoning traces. 
We further introduce the Chain-of-Checklist (CoCL) reasoning paradigm and adopt a two-stage training strategy to achieve more reliable and comprehensive error detection. 
Extensive experiments on \benchopt show that \modelopt-7B attains state-of-the-art performance in parsing quality evaluation. 
Additionally, its detailed quality assessment guides model refinement effectively, demonstrating both its reliability and exceptional performance as an evaluator and its potential to advance the development of document parsing.

{
    \small
    \bibliographystyle{ieeenat_fullname}
    \bibliography{main}
}

\newpage

\clearpage
\setcounter{page}{1}
\maketitlesupplementary

\setcounter{section}{0}
\setcounter{table}{0}
\setcounter{figure}{0}
\renewcommand{\thesection}{\Roman{section}}
\renewcommand{\thetable}{S\arabic{table}}
\renewcommand{\thefigure}{S\arabic{figure}}

\textbf{Contents of the Appendices}\\
\Rmnum{1}. Definition of error types.\\
\Rmnum{2}. Construction Details of \datasetopt\\
\Rmnum{3}. Composition Details of \datasetopt\\
\Rmnum{4}. Template of Chain-of-Checklist(CoCL)\\
\Rmnum{5}. Details of \benchopt\\
\Rmnum{6}. More Training Details\\
\Rmnum{7}. Prompt of Experiments\\
\Rmnum{8}. Case Study\\


\section{Definition of Error Types}
\label{appendix:error_type_definition}
\begin{figure*}[h]
  \centering
   \includegraphics[width=1.0\linewidth]{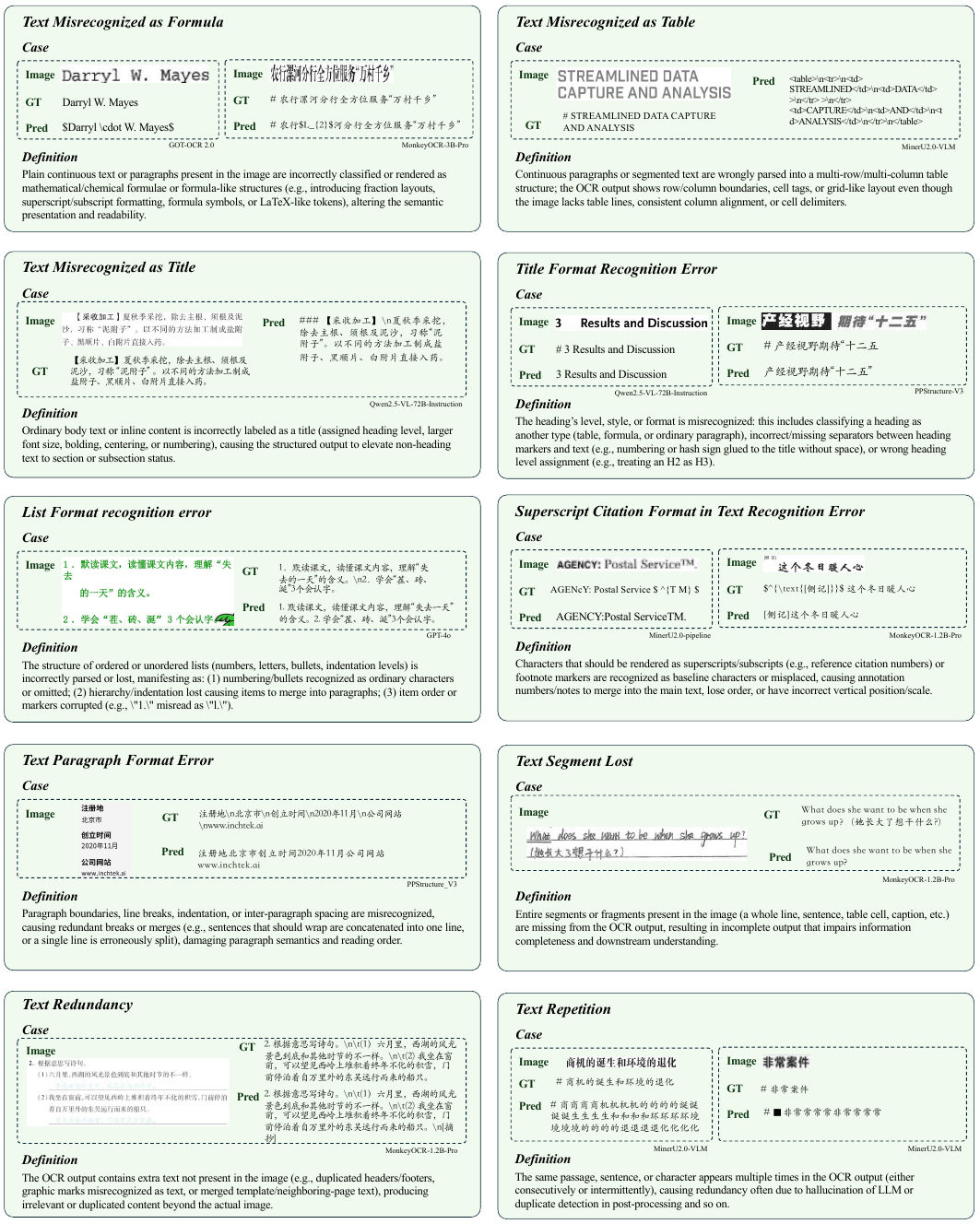}
    \vspace{-6mm}
   \caption{Definition and instance of error type(1).}
   \label{fig:app_definition_1}
   \vspace{-6mm}
\end{figure*}

\begin{figure*}[h]
  \centering
   \includegraphics[width=1.0\linewidth]{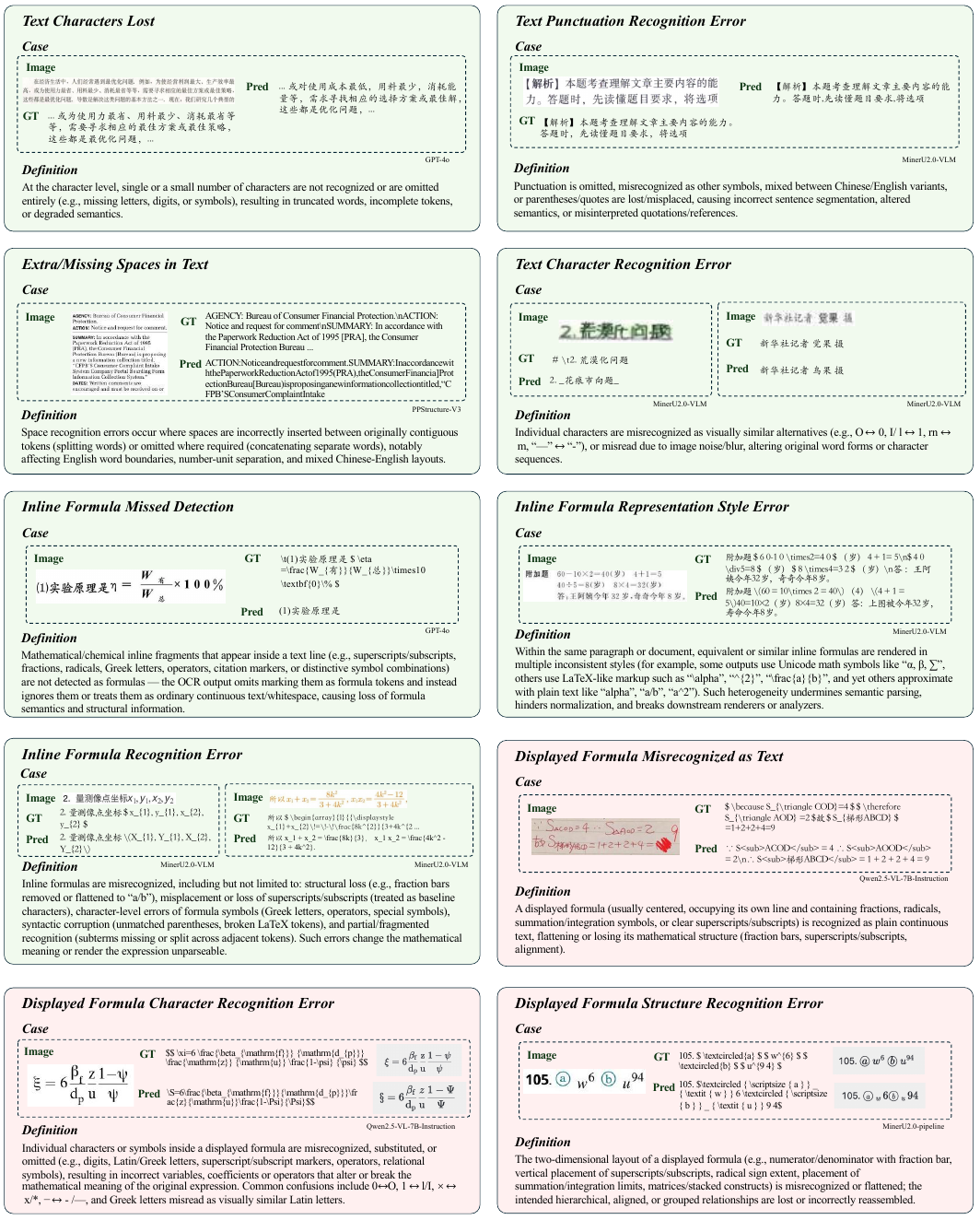}
    \vspace{-6mm}
   \caption{Definition and instance of error type(2).}
   \label{fig:app_definition_2}
   \vspace{-6mm}
\end{figure*}

\begin{figure*}[h]
  \centering
   \includegraphics[width=1.0\linewidth]{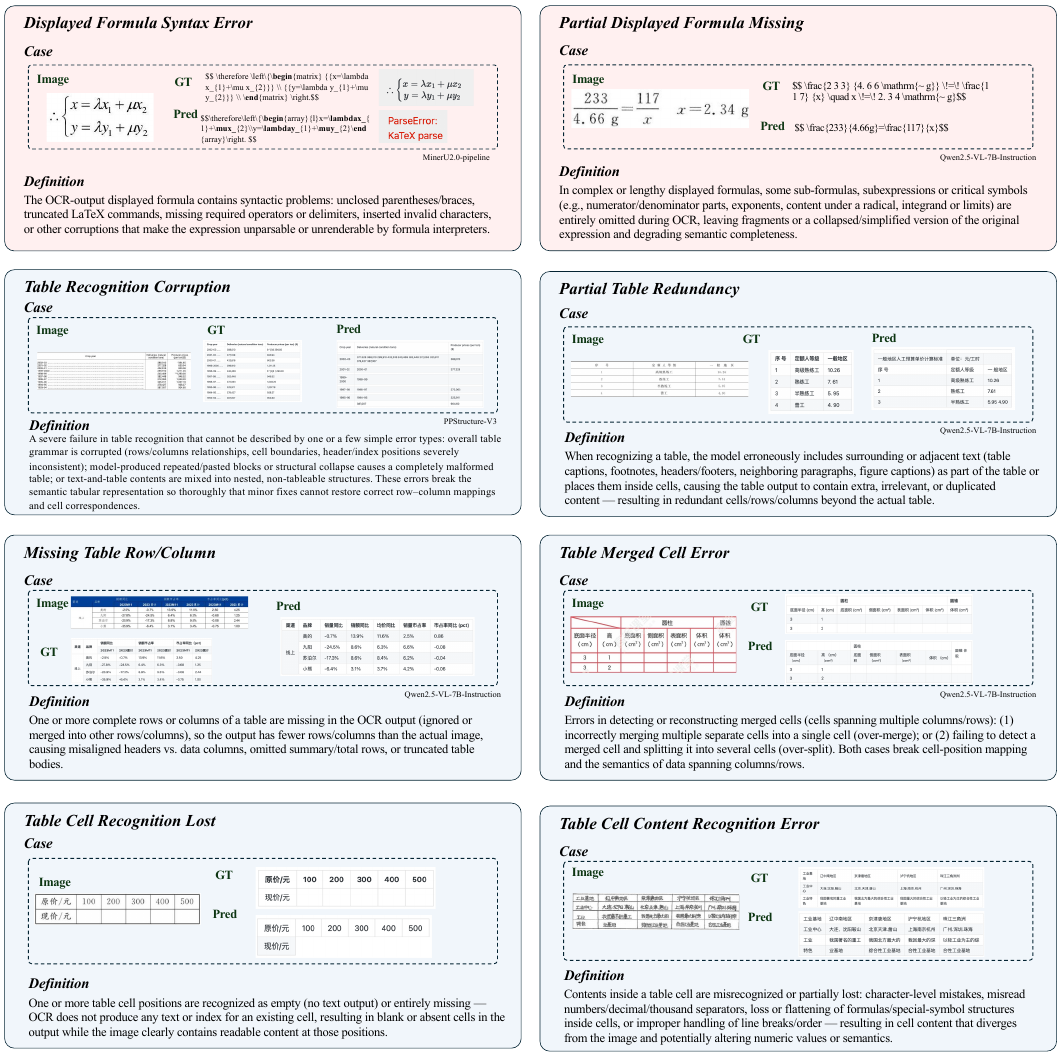}
    \vspace{-6mm}
   \caption{Definition and instance of error type(3).}
   \label{fig:app_definition_3}
   \vspace{-6mm}
\end{figure*}

In this section, we provide detailed definitions of the 28 error types and present corresponding case studies.
As shown in Figures~\cref{fig:app_definition_1},~\cref{fig:app_definition_2}, and~\cref{fig:app_definition_3}, we illustrate the definition of each error type along with 1–2 real-world examples in \benchopt collected from document parsing models.

\clearpage
\section{Construction Details of \datasetopt}
\label{appendix:dataset}

\subsection{Error Type Synthesis Scheme}
\label{appendix:error_case_gen}
This section details the synthesis methodology for each error type. For rule-based error categories, we elaborate on the synthesis logic and implementation procedures. For LLM-guided error types, we provide complete prompt templates and employ few-shot learning to enhance generation quality. All synthetic data is produced using Gemini-2.5-Flash.

\subsubsection{Text Error Type Synthesis Scheme}
\label{appendix:error_case_gen_text}
Among the 14 text error types, 11 of them are synthesized using rule-based methods, while the remaining 3 employ LLM-based synthesis. Detailed prompt templates for text error synthesis are documented in Table~\cref{tab:prompt_text}.
For the three inline formula-related error types in text: \textit{Inline Formula Missed Detection} utilizes a rule-based approach; \textit{Inline Formula Representation Style Error} employs a custom-designed prompt (see~\cref{tab:prompt_inline_formula}); and the remaining type directly adopts the LLM-based synthesis scheme for equation errors outlined in~\cref{tab:prompt_equation_1} and~\cref{tab:prompt_equation_2}.

\noindent\textbf{Type Level Error of Text}
\begin{itemize}
    \item \textbf{Text misrecognized as table}: uses LLM-guided synthesis, adding HTML table syntax to single-line or multi-line well-structured text.
    \item \textbf{Text misrecognized as formula}: uses LLM-guided synthesis, adding \LaTeX{} math syntax to short text segments.
    \item \textbf{Text misrecognized as title}: uses rule-based synthesis by randomly adding 1–3 ``\#'' characters to the beginning of short text.
\end{itemize}

\noindent\textbf{Format Level Error of Text}
\begin{itemize}
    \item \textbf{Text paragraph format error}: uses rule-based synthesis by randomly deleting (1 to max) or inserting \verb|\|n (1–5 times) inside a paragraph.
    \item \textbf{List format recognition error}: uses rule-based synthesis by perturbing existing list formats, such as deleting the \verb|\|n after list indices.
    \item \textbf{Title format recognition error}: uses rule-based synthesis by removing heading syntax and converting the title to plain text.
    \item \textbf{Superscript citation format in text recognition error}: uses rule-based synthesis to detect superscript/subscript structures and Unicode superscript/subscript characters in non-formula regions, replacing them with plain-text equivalents.
\end{itemize}

\noindent\textbf{Span Level Error of Text}
\begin{itemize}
    \item \textbf{Text repetition}: uses rule-based synthesis by selecting a random tail segment or a span between two punctuation marks and repeating it 10–20 times.
    \item \textbf{Text redundancy}: uses rule-based synthesis by randomly inserting text from other samples into paragraphs, formulas, or tables.
    \item \textbf{Text segment lost}: uses rule-based synthesis by randomly deleting content between any two punctuation marks.
\end{itemize}

\noindent\textbf{Character Level Error of Text}
\begin{itemize}
    \item \textbf{Text characters lost}: uses rule-based synthesis by randomly deleting 1–5 Chinese characters, or 1–3 English words, or 1–5 English characters.
    \item \textbf{Text punctuation recognition error}: uses rule-based synthesis by randomly applying one perturbation strategy: deleting arbitrary punctuation, deleting paired punctuation, or swapping between Chinese and English punctuation.
    \item \textbf{Extra/missing spaces in text}: uses rule-based synthesis by randomly deleting or inserting spaces.
    \item \textbf{Text character recognition error}: uses LLM-guided synthesis to simulate both glyph-similarity-based and semantic-similarity-based character recognition errors.
\end{itemize}

\noindent\textbf{Inline-Formula Error of Text}
\begin{itemize}
    \item \textbf{Inline formula missed detection}: uses rule-based synthesis by detecting existing inline formulas in text and randomly deleting any number of them.
    \item \textbf{Inline formula recognition error}: uses rule-based synthesis by injecting one randomly selected formula error—syntax error, structural error, character error, or partial omission—into relatively long inline formulas.
    \item \textbf{Inline formula style error}: uses LLM-guided synthesis by rewriting a subset of inline formulas within multi-formula text blocks into their Unicode-style representations.
\end{itemize}

\subsubsection{Table Error Type Synthesis Scheme}
\label{appendix:error_case_gen_table}
Among the six table error types, \textit{Missing Table Row/Column} and \textit{Table Cell Recognition Loss} employ rule-based synthesis methods. For \textit{Table Recognition Corruption}, we utilize an LLM to filter instances from real-world samples (see~\cref{tab:table_prompt_table_fliter}). The remaining three error types are synthesized using LLM-guided approaches, with detailed prompts provided in~\cref{tab:prompt_table}.

\noindent\textbf{Integrity Level Error of Table}
\begin{itemize}
    \item \textbf{Partial table redundancy}: uses LLM-guided synthesis by adding redundant rows before or after the table.
    \item \textbf{Table recognition corruption}: based on the Real-world Case Selection method, models such as Qwen2.5-VL-7B and InternVL3-8B are sampled multiple times with temperature=1.0, and low-quality samples are further filtered by an LLM.
\end{itemize}

\noindent\textbf{Structure Level Error of Table}
\begin{itemize}
    \item \textbf{Missing table row/column}: uses rule-based methods by randomly deleting rows or columns (potentially deleting both simultaneously).
    \item \textbf{Table merged cell error}: uses LLM-guided synthesis by randomly merging table cells or modifying existing merged cells.
    \item \textbf{Table cell recognition lost}: uses rule-based methods by randomly deleting a certain number of table cells.
\end{itemize}

\noindent\textbf{Content Level Error of Table}
\begin{itemize}
    \item \textbf{Table cell content recognition error}: uses LLM-guided synthesis by injecting recognition errors into the content of randomly selected table cells.
\end{itemize}

\subsubsection{Equation Error Type Synthesis Scheme}
Given the inherent syntactic and structural complexity of mathematical expressions, all five equation error types are synthesized using an LLM-guided approach. Detailed prompt templates are provided in~\cref{tab:prompt_equation_1} and ~\cref{tab:prompt_equation_2}.

\noindent\textbf{Type Level Error of Equation}
\begin{enumerate}
    \item \textbf{Displayed formula misrecognized as text}: uses LLM-guided synthesis by converting formatted equations in LaTeX into Unicode representations or plain text.
\end{enumerate}

\noindent\textbf{Integrity Level Error of Equation}
\begin{enumerate}
    \item \textbf{Displayed formula syntax error}: uses LLM-guided synthesis by injecting non-standard or incorrect LaTeX syntax into formulas.
    \item \textbf{Partial displayed formula missing}: uses LLM-guided synthesis by randomly deleting sub-expressions from long formulas.
\end{enumerate}

\noindent\textbf{Content Level Error of Equation}
\begin{enumerate}
    \item \textbf{Displayed formula structure recognition error}: uses LLM-guided synthesis by randomly modifying structural control symbols or directly altering formula structures.
    \item \textbf{Displayed formula character recognition error}: uses LLM-guided synthesis by randomly modifying characters within formulas.
\end{enumerate}

\begin{table*}[t]
    \centering
    \small
    \begin{tabular}{p{0.9\linewidth}}
    \toprule
    \textit{Prompt Tamplate} \\
    \midrule
    \textbf{Role} \\
    \quad You are an expert in simulating optical character recognition (OCR) errors. I will provide you with plain text. Your task is to simulate a common OCR error: \textcolor{blue}{\{detailed\_task\_description\}}.\\
    \textbf{Task Guidelines} \\
    \quad\textcolor{blue}{\{specific\_task\_instructions\}}\\
    \textbf{Output Formats} \\
    \quad \textcolor{blue}{\{output\_formats\}} \\
    \textbf{Examples} \\
    \quad \textcolor{blue}{\{example\_list\}} \\
    \quad Now, let's get started! Please modify the following text:\\
    \midrule
    \textit{Text misrecognized as table} \\
    \midrule
    \sethlcolor{yellow}\hl{\textbf{detailed task description:}} \\
    \quad **Misrecognizing continuous text as a table**. You should **rewrite the input as an HTML table** to simulate that OCR may incorrectly split and align the text into rows and columns.\\
    \sethlcolor{yellow}\hl{\textbf{specific task instructions:}} \\
    \quad 1. When the text is short, you can rewrite it as a simple structure such as a single-row and single-column table.\\
    \quad 2. You may incorrectly split sentences into table cells.\\
    \quad 3. You may misplace words between rows or columns.\\
    \quad  4. You can insert $<td>$ to simulate an illusion grid structure.\\
    \sethlcolor{yellow}\hl{\textbf{Output Format:}} \\
    \quad Original Text: [Input Text] \\
    \quad Final Text: [HTML Table Simulating OCR Error]\\
    \midrule
    \textit{Text misrecognized as formula} \\
    \midrule
    \sethlcolor{yellow}\hl{\textbf{detailed task description:}} \\
    \quad **plain text gets misrecognized as mathematical formulas**. This type of error is common in titles in images, short text in bold, and the model is prone to recognize such text as a formula format due to style confusion.\\
    \sethlcolor{yellow}\hl{\textbf{specific task instructions:}} \\
    \quad 1. Wrap text segments with formula delimiters (\$...\$ or \$\$...\$
    \$)\\
    \quad 2. For short texts: perform complete formula conversion; For long texts: selectively convert portions. \\
    \quad 3. You can simulate error patterns, such as: \\
    \quad \quad - Misinterpret continuous letters as variable products.\\
    \quad \quad - Misidentify punctuation as mathematical symbols.\\
    \sethlcolor{yellow}\hl{\textbf{Output Format:}} \\
    \quad Original Text: [Input Text] \\
    \quad Final Text: [Latex Formula Simulating OCR Error]\\
    \midrule
    \textit{Text character recognition error} \\
    \midrule
    \sethlcolor{yellow}\hl{\textbf{detailed task description:}} \\
    \quad **text character recognition error**. You should modify 1 to 5 characters within the text to mimic common OCR mistakes.\\
    \sethlcolor{yellow}\hl{\textbf{specific task instructions:}} \\
    \quad 1. Visually Similar Characters: Replace characters with visually confusable ones (e.g., "O" → "0", "l" → "1"). \\
    \quad 2.Semantic Hallucination: Substitute characters/words that might arise from contextual misprediction (e.g., "apple" → "apricot"). \\
    \sethlcolor{yellow}\hl{\textbf{Output Format:}} \\
    \quad Original Text: [Input Text] \\
    \quad Total changes: [numbers] \\
    \quad Modification Details:  \\
    \quad \quad 1. [original char] → [new char] (Position: [index])  \\
    \quad \quad 2. ...  \\
    \quad Final Text: [modified text]  \\
    \bottomrule
    \end{tabular}
    \caption{Text error case Generation Prompt}
    \label{tab:prompt_text}
\end{table*}

\begin{table*}[t]
    \centering
    \begin{tabular}{p{0.9\linewidth}}
    \toprule
    \textit{Inline formula representation style error} \\
    \midrule
    \textbf{role} \\
    \quad You are an expert in simulating structural syntax errors in formula recognition (e.g. OCR, handwriting conversion). Your task is to traverse a given LaTeX formula and determine whether it meets the requirements. If it meets the requirements, convert it to a formula represented by unicode to simulate the scenario in which the formula structure content may be output as unicode special characters in the real formula recognition task.\\
    \textbf{Task Guidelines} \\
    \quad Required rules need to meet at least one of the following:\\
    \quad\quad 1. Structures represented by unicode can be used, such as formula structures such as subscripts and subscripts, such as $a^2$, $km^2$, etc.;\\
    \quad \quad 2. In unicode, there are corresponding special characters, such as Greek letters $ \alpha $, operation symbols, etc.;\\
    If the above requirements are met, rewrite this formula to use unicode. Otherwise, output the original formula.\\
    \quad Please note:\\
    \quad \quad 1. In addition to converting the format and special characters to Unicode, if the other contents in this formula can be accurately expressed in ordinary text, there is no need to change them, for example, "2" can be written as "2".\\
    \quad \quad 2. The modified formula does not need to be wrapped in formula symbols.\\
    \textbf{Output Formats} \\
    \quad Output format: \\
    \quad Original formula: [Input]\\
    \quad Modification details:\\
    \quad\quad 1. [Original sub-formula] → [Modified sub-formula]\\
    \quad\quad 2. ...\\
    \quad Final formula: [Modified formula]\\
    \textbf{Examples} \\
    \quad \textbf{Example 1:}\\
    \quad Input: \(\frac{a + b}{2} = c^{2}\)\\
    \quad Output:\\
    \quad \quad Original formula: \(\frac{a + b}{2} = c^{2}\)\\
    \quad \quad Modification details:\\
    \quad \quad 1. "\text{\textbackslash frac\{a + b\}\{2\}}" → "(a+b)/2" \\
    \quad \quad 2. "c\^\{2\}" → "c²" \\
    \quad \quad Final formula: (a+b)/2 = c² \\
    \quad \textbf{Example 2:}\\
    \quad Input: \(27.5 \%\)\\
    \quad Output:\\
    \quad\quad Original formula: \(27.5 \%\)\\
    \quad\quad Modification details:\\
    \quad\quad This formula does not contain unicdoe, no need to use unicode modification\\
    \quad\quad Final formula: \(27.5 \%\)\\
    \\
    \quad Now, let's get started! Please modify the following text:\\
    \bottomrule
    \end{tabular}
    \caption{Case with Inline Formula Representation Style Error Generation Prompt}
    \label{tab:prompt_inline_formula}
\end{table*}

\begin{table*}[t]
    \centering
    \footnotesize
    \begin{tabular}{p{0.9\linewidth}}
    \toprule
    \textit{Prompt Tamplate} \\
    \midrule
    \textbf{Role} \\
    \quad You are a professional expert in simulating table recognition errors, specializing in replicating various types of errors that occur during OCR recognition, data parsing, and format conversion.  I will provide you with a table in HTML. Your task is to a common table recognition error: \textcolor{blue}{\{detailed\_task\_description\}}.\\
    \textbf{Task Guidelines} \\
    \quad\textcolor{blue}{\{specific\_task\_instructions\}}\\
    \textbf{Output Formats} \\
    \quad Original Table: [Tale in HTML]\\
    \quad Modification Details: \\ 
    \quad\quad 1.  [Original content] → [Modified content] (Error Type)  \\
    \quad\quad 2. ... \\
    \quad Final Table: [Modified Table in HTML]\\
    \textbf{Examples} \\
    \quad \textcolor{blue}{\{example\_list\}} \\
    \quad Now, let's get started! Please modify the following table:\\
    \midrule
    \textit{Partial table redundancy} \\
    \midrule
    \sethlcolor{yellow}\hl{\textbf{detailed task description:}} \\
    \quad inject **partial redundancy errors** into a given HTML table by modifying the HTML content.\\
    \sethlcolor{yellow}\hl{\textbf{specific task instructions:}} \\
    \quad1. Pre-/Post-table Redundancy:  \\
    \quad\quad Add extra rows **before** or **after** the main table content.  \\
    \quad\quad These rows simulate the case where the surrounding text is mistakenly recognized as part of the table.  \\
    \quad\quad - You may generate synthetic content such as "Table begins", "Data continues", etc. \\
    \quad 2. Internal Redundant Content:  \\
    \quad\quad Insert redundant or duplicate content **inside** the table. You may:\\
    \quad\quad- Add a row with duplicated or empty cells\\
    \quad\quad- Insert a column with similar or empty cells\\
    \quad\quad- Duplicate an existing cell and place it again in the same row or another row\\
    \midrule
    \textit{Table merged cell error} \\
    \midrule
    \sethlcolor{yellow}\hl{\textbf{detailed task description:}} \\
    \quad simulate **merged cell errors (e.g., issues related to `colspan` and `rowspan`)** commonly seen in real OCR scenarios, modifying the original HTML table while keeping its structure basically identifiable \\
    \sethlcolor{yellow}\hl{\textbf{specific task instructions:}} \\
    \quad 1. **Disrupting Existing Merged Structures**  \\
    \quad If the original table contains `colspan` or `rowspan`, you may:  \\
    \quad\quad - Alter their values to mismatch the actual layout   \\
    \quad\quad - Remove these attributes entirely \\
    \quad\quad - Create span overlaps or inconsistencies across rows/columns \\
    \quad 2. **Introducing Incorrect Merging**  \\
    \quad\quad If the original table has no `colspan`/`rowspan`, you may randomly insert these attributes with incorrect values to simulate hallucinated or misrecognized merges.\\
    \midrule
    \textit{Table cell content recognition error} \\
    \midrule
    \sethlcolor{yellow}\hl{\textbf{detailed task description:}} \\
    \quad simulate errors in the content of an HTML table **while strictly preserving the overall table structure**. \\
    \sethlcolor{yellow}\hl{\textbf{specific task instructions:}} \\
    \quad 1. Structure Preservation  \\
    \quad Strictly maintain the original table’s tag structure, attributes, and overall layout.  \\
    \quad 2. Error Types (Choose the most context-appropriate 1–3 types)  \\
    \quad - Visual Confusion: Replace visually similar characters (case-sensitive, full-width/half-width), including symbols, numbers, letters and Chinese characters. \\
    \quad - Whitespace Errors, including removing all spaces, adding extra spaces randomly, mixing of English and Chinese space characters \\
    \quad - Format Loss, including remove formulas (e.g., "$c = {a}_{b}$" → "c = ab") and flattening list content (e.g., "- Item1$\setminus$n- Item2" → "- Item1 - Item2") \\
    \quad - Punctuation Errors, including swapping between Chinese and English punctuation and incorrecting recognition of punctuation marks \\ 
    \quad 3. Modification Intensity: Large tables ($>$20 cells): Modify 3–5 places; Medium tables (5–20 cells): Modify 2–3 places; Small tables ($<$5 cells): Modify 1 place \\ 
    \bottomrule
    \end{tabular}
    \caption{Table Error Case Generation Prompt}
    \label{tab:prompt_table}
\end{table*}

\begin{table*}[t]
    \centering
    \begin{tabular}{p{0.9\linewidth}}
    \toprule
    \textit{Table recognition corruption} \\
    \midrule
    \textbf{role} \\
    \quad You are an expert in table recognition and evaluation. Your task is to assess the quality of an HTML table generated by a model by comparing it with a given ground-truth (GT) table. Follow the instructions below strictly:\\
    \textbf{Task Guidelines} \\
    \quad 1. Check Renderability\\
    \quad - Determine whether the predicted table can be successfully rendered as HTML.\\
    \quad - Mark as unrenderable if it contains:\\
    \quad \quad - Syntax errors (e.g., missing or mismatched tags)\\
    \quad \quad - Repeated or nested table structures that cause rendering failure\\
    \quad - Non-HTML formats (e.g., LaTeX, Markdown)\\
    \quad 2. Check Structural Disorder\\
    \quad Judge whether the predicted table is seriously disordered compared to the GT. Consider a table seriously disordered if: \\
    \quad\quad The difference between prediction and GT cannot be simply described as one or more of: \\
    \quad\quad 1. A small number of wrong cell contents \\
    \quad\quad 2. A few missing or extra cells \\
    \quad\quad 3. A few incorrect or missing merged cells \\
    \quad Instead, it shows overall structural mismatch, misaligned rows/columns, or chaotic layout.
    \quad Format Handling\\
    \quad If the predicted table is not in HTML format (e.g., LaTeX, Markdown), return: Unable to judge \\
    \\
    \textbf{Final Judgment}\\
    \quad Based on your analysis, output either: \\
    \quad\quad - Bad Table if the table cannot be rendered or is seriously disordered \\
    \quad\quad - Good Table if the structure is mostly aligned and valid \\
    \textbf{Output Formats} \\
    \quad Output format: \\
    \quad [Analysis process]\\
    \quad Your detailed reasoning here: e.g., syntax check, structure check, comparison to GT...\\
    \quad\quad [Result] Bad Table / Good Table / Unable to judge\\
    \quad Now, let's get started! Please modify the following text:\\
    \bottomrule
    \end{tabular}
    \caption{Judge table recognition corruption Prompt}
    \label{tab:table_prompt_table_fliter}
\end{table*}   

\begin{table*}[t]
    \centering
    \small
    \begin{tabular}{p{0.9\linewidth}}
    \toprule
    \textit{Equation Prompt Tamplate} \\
    \midrule
    \textbf{Role} \\
    \quad You are an expert in simulating formula recognition errors. I will provide you with a formula in LaTeX. Your task is to: \textcolor{blue}{\{detailed\_task\_description\}}.\\
    \textbf{Task Guidelines} \\
    \quad\textcolor{blue}{\{specific\_task\_instructions\}}\\
    \textbf{Output Formats} \\
    \quad \textcolor{blue}{\{output\_formats\}} \\
    \textbf{Examples} \\
    \quad \textcolor{blue}{\{example\_list\}} \\
    \quad Now, let's get started! Please modify the following formula:\\
    \midrule
    \textit{Displayed formula misrecognized as text} \\
    \midrule
    \sethlcolor{yellow}\hl{\textbf{detailed task description:}} \\
    \quad simulate **Displayed formula misrecognized as text** error. You should convert the given LaTeX mathematical formulas into plain text or Unicode representations, simulating scenarios where formulas are incorrectly not expressed in LaTeX during real-world formula recognition tasks.\\
    \sethlcolor{yellow}\hl{\textbf{specific task instructions:}} \\
    \quad 1. Express formulas using plain text whenever possible, without wrapping them in formula symbols;\\
    \quad 2. Represent special characters, Greek letters, and operators from LaTeX using their Unicode equivalents;\\
    \quad 3. It's acceptable if complex formula structures cannot be perfectly represented with text and Unicode.\\
    \sethlcolor{yellow}\hl{\textbf{Output Format:}} \\
    \quad Original formula: [Input] \\
    \quad Final formula: [Modified Formula]\\
    \midrule
    \textit{Partial displayed formula missing} \\
    \midrule
    \sethlcolor{yellow}\hl{\textbf{detailed task description:}} \\
    \quad simulate **Partial displayed formula missing** error. You should randomly delete a sub-formula part in this formula to simulate the common situation of missing parts in formula recognition.\\
    \sethlcolor{yellow}\hl{\textbf{specific task instructions:}} \\
    \quad 1. The final formula also needs to be correctly wrapped with latex symbols, such as "$\setminus$(" and "$\setminus$)", "$\setminus$[" and "$\setminus$]", etc.;\\
    \quad 2. The deleted sub-formula should not be too long;\\
    \sethlcolor{yellow}\hl{\textbf{Output Format:}} \\
    \quad Original text: [Input formula] \\
    \quad Deleted sub-formula part: [Part of the formula content]\\
    \quad Final formula: [Final formula]\\
    \midrule
    \textit{Displayed formula syntax error} \\
    \midrule
    \sethlcolor{yellow}\hl{\textbf{detailed task description:}} \\
    \quad simulate **Displayed formula syntax error** error. \\
    \sethlcolor{yellow}\hl{\textbf{specific task instructions:}} \\
    \quad 1. You should severely perturb this formula to simulate the situation in which the model cannot recognize correctly due to the lack of model ability, such as serious syntax errors and repeated generation.
    \quad 2. Please note that the final formula also needs to be correctly wrapped with latex symbols, such as "$\setminus$(" and "$\setminus$)", "$\setminus$[" and "$\setminus$]", etc\\
    \sethlcolor{yellow}\hl{\textbf{Output Format:}} \\
    \quad Original text: [Input formula] \\
    \quad Final formula: [Modified formula]\\
    \bottomrule
    \end{tabular}
    \caption{Equation error case Generation Prompt(2)}
    \label{tab:prompt_equation_1}
\end{table*}   

\begin{table*}[t]
    \centering
    \small
    \begin{tabular}{p{0.9\linewidth}}
    \toprule
    \textit{Displayed formula structure recognition error} \\
    \midrule
    \sethlcolor{yellow}\hl{\textbf{detailed task description:}} \\
    \quad simulate **Displayed formula structure recognition error** error. You should make 1-5 random structural modifications to LaTeX formulas according to the following rules to simulate possible structural errors in real formula recognition tasks.\\
    \sethlcolor{yellow}\hl{\textbf{specific task instructions:}} \\
    \quad 1. Only structural control symbols and characters involved in structural control symbols can be modified: Fractions($\frac{}{}$ → $\setminus$ or $\over$), Exponents/subscripts, Roots: ($\sqrt{ab} + c$ → $\sqrt{ab + c}$), Matrix(swap the positions of elements in the matrix), Brackets(delete brackets such as () and {} as appropriate), Operators($\hat{}$ → $\vec{}$). And not limited to the above rules.\\
    \quad 2. Modification is prohibited: Text characters (e.g., a, 2, x), Semantic meaning (e.g., do not modify the representation of Greek letters, do not change $\setminus$alpha to a or unicode representation).
    \quad 3. Confusion is allowed: Visually similar symbols (e.g., $\setminus$frac → $\setminus$tfrac), Structural deletion (e.g., delete { } around exponents).
    \quad 4. Keep LaTeX brackets (keep $\setminus$( $\setminus$) or $\setminus$[ $\setminus$]). The final result should not have too many syntax errors, and it needs to be rendered successfully.\\
    \sethlcolor{yellow}\hl{\textbf{Output Format:}} \\
    \quad Original text: [Input formula] \\
    \quad Number of modifications: [1-5] \\
    \quad Modification details: \\
    \quad \quad 1. [original] → [modified] \\
    \quad \quad 2. ... \\
    \quad Final formula: [Modified formula]\\
    \midrule
    \textit{Displayed formula character recognition errorr} \\
    \midrule
    \sethlcolor{yellow}\hl{\textbf{detailed task description:}} \\
    \quad simulate **Displayed formula character recognition error** error. You should randomly modify 1 to 5 characters in the formula to simulate common content recognition errors in formula recognition.\\
    \sethlcolor{yellow}\hl{\textbf{specific task instructions:}} \\
    \quad 1. Only modify the characters and numbers in the formula, not the content related to the formula structure;\\
    \quad 2. You can replace real characters with visually confusing characters (such as "O" → "0", "l" → "1"); swap the uppercase and lowercase letters of English letters and Greek letters; \\
    \quad 3. You can interchange visually similar English letters with Greek letters, as well as interchange the upper and lower case of English letters and Greek letters;\\
    \quad 4. You can modify the style of the characters in the formula, such as adding, modifying, deleting bold characters, etc.\\
    \quad 5. The final formula also needs to be correctly wrapped with latex symbols, such as "$\setminus$(" and "$\setminus$)", "$\setminus$[" and "$\setminus$]", etc.
    \sethlcolor{yellow}\hl{\textbf{Output Format:}} \\
    \quad Original text: [Input formula] \\
    \quad Number of modifications: [1-5] \\
    \quad Modification details: \\
    \quad \quad 1. 1. [Original character] → [Modified character] \\
    \quad \quad 2. ... \\
    \quad Final formula: [Modified formula]\\
    \bottomrule
    \end{tabular}
    \caption{Equation error case Generation Prompt(2)}
    \label{tab:prompt_equation_2}
\end{table*}

\subsection{Reasoning Generation}
We design structured prompt templates and employ a few-shot strategy to guide the generation of reasoning chains individually for each error type as shown in ~\cref{tab:reason_prompt}. For every document element instance, we provide the image, reference ground truth, and the parsed output containing specific error types to Qwen2.5-VL-72B-Instruct, directing it to produce dedicated reasoning processes for each error type.

\begin{table*}[t]
    \centering
    \small
    \begin{tabular}{p{0.9\linewidth}}
    \toprule
    \textit{Reason Construction Prompt Tamplate} \\
    \midrule
    \textbf{Role} \\
    \quad You are a professional document layout analyst and OCR error-validation expert. You will receive a document image, a reference OCR result, and another OCR result that contains an \textcolor{blue}{\{error\_type\}} error.\\
    \textbf{Definition of error} \\
    The definition of \textcolor{blue}{\{error\_type\}} is: \textcolor{blue}{\{error\_definition\}}\\
    \textbf{Input} \\
    Inputs (you will receive three items):\\
    \quad  1) Document image —— a crop of the document image; \\
    \quad 2) Reference OCR result of the image;\\
    \quad 3) OCR result with \textcolor{blue}{\{error\_type\}} error from an OCR model;\\
    \textbf{Task Guidelines} \\
    \quad  - Compare the visual information in the image with the OCR outputs and, based on observable visual/structural evidence, produce 1–2 concise sentences in English explaining why the OCR result was judged to have the \textcolor{blue}{\{error\_type\}}.\\
    \quad  - Analyze only the specified error type (ignore other OCR issues even if present). \\
    \quad  - Output must contain only the 1–2 sentence explanation in English — no headings, labels, input repetition, examples, or extra commentary. \\
    \quad  - Use assertive, evidence-based wording; avoid hedging expressions like "maybe" or "possibly."\\
    \\
    \quad Now, let's begin!\\
    \quad Input: \\
    \quad\quad IMAGE: $<$image$>$\\
    \quad\quad REFERENCE\_OCR: \textcolor{blue}{\{reference\_ocr\}} \\
    \quad\quad OCR\_RESULT\_WITH\_ERROR: \textcolor{blue}{\{ocr\_result\}}\\
    \quad\quad ERROR\_TYPE: \textcolor{blue}{\{error\_type\}}\\
    \textbf{Examples} \\
    \quad \textcolor{blue}{\{golden\_answer\}
    } \\
    \midrule
    \textit{Detailed description} \\
    \midrule
    \sethlcolor{yellow}\hl{\textbf{error\_type:}} \\
    One of the 28 specific error types.\\
    \sethlcolor{yellow}\hl{\textbf{error\_definition:}} \\
    Definitions provided in~\cref{fig:app_definition_1},~\cref{fig:app_definition_2} and~\cref{fig:app_definition_3}.\\
    \sethlcolor{yellow}\hl{\textbf{golden\_answer:}} \\
    2-3 manually crafted reasoning demonstrations for this specific error type.\\
    \bottomrule
    \end{tabular}
    \caption{Prompt for reason construction of each error type in case }
    \label{tab:reason_prompt}
\end{table*}

\clearpage

\section{Composition Details of \datasetopt}
\textbf{Crop Image Distribution}. The \datasetopt comprises 84,176 cropped element images. Purely English content accounts for 53.60\%, with the remaining portion containing Chinese characters. The three major element types—Text, Table and Equation-account for 65.19\%, 17.82\%, and 16.98\% of the total images, respectively.

\noindent\textbf{Detailed Element Case Distribution}. In Table~\ref{tab:case_type}, we present the distribution of the 11 fine-grained element case types within \benchopt.

\begin{table}[h]
  \centering
  \caption{Statistics of 11 classes of element cases in \benchopt}
  \label{tab:case_type}
  \small
  \begin{tabular}{l c c}
    \toprule 
    \textbf{Element} & \textbf{Count} & \textbf{Percentage} \\
    \midrule 
    Text Block & 45868 & 21.59\%\\
    Text with inline-formula & 45791 & 21.56\%\\
    Title & 8996 & 4.23\%\\
    List & 5992 & 2.82\%\\
    Image Caption & 2498 & 1.18\%\\
    Table Caption & 749 & 0.35\%\\
    Table Footnote & 112 & 0.05\%\\
    Table & 57510 & 27.07\%\\
    Equation& 44874 & 21.12\%\\
    \bottomrule 
  \end{tabular}
\end{table}

\noindent\textbf{Error Count Distribution}. Table~\ref{tab:error_nums} shows the proportion of cases containing different numbers of error types. Cases without any errors are referred to as \textit{Good Cases}, whereas those containing single or multiple types of errors are uniformly classified as \textit{Bad Cases}. 
\begin{table}[h]
  \centering
  \caption{Statistics of cases containing different numbers of error types in \datasetopt}
  \small
  \begin{tabular}{c c c c c}
    \toprule
    & & \textbf{Text} & \textbf{Table} & \textbf{Formula} \\
    \midrule
    \multicolumn{2}{c}{\textbf{Good Case}} & 20003 & 10000 & 10000 \\
    \multirow{3}{*}{\textbf{Bad Case}} & 1 error type & 74848 & 37779 & 25810 \\
     & 2 error types & 6712 & 6851 & 6289 \\
     & 3 error types & 7554 & 2880 & 2518 \\
    \multicolumn{2}{c}{\textbf{Total}}  & 109117 & 57510 & 44617 \\
    \bottomrule
  \end{tabular}
  \label{tab:error_nums}
\end{table}

\noindent\textbf{Error Type Distribution}. In Table~\ref{tab:error_type}, we specifically present the distribution of the different error types.

\newpage
\section{Details of \benchopt}
\label{appendix:benchmark}
\noindent\textbf{Data Sources}. In \modelopt, 92.7\% of the samples are real, originating from various sources including MinerU2.0-pipeline~\cite{wang2024mineru}, PP-StructureV3~\cite{cui2025paddleocr}, GPT-4o~\cite{GPT-4o}, Qwen2.5-VL-7B-Instrcut~\cite{bai2025qwen2}, MonkeyOCR-1.2B-Pro~\cite{li2025monkeyocr}, and MinerU2.0-vlm~\cite{wang2024mineru}. To ensure a balanced error type distribution, 7.3\% of the cases were manually synthesized by experts. The detailed statistics of the source distribution are presented in Table~\ref{tab:bench_model}. 

\begin{table}[h]
    \centering
    \caption{Statistics of error type in \benchopt}
    \label{tab:bench_model}
    \small
    \begin{tabular}{lcc}
        \toprule
        \textbf{Model} & \textbf{Count} & \textbf{Percentage} \\
        \midrule
        MonkeyOCR-1.2B-Pro & 142 & 16.1\% \\
        PP-StructureV3 & 99 & 11.2\% \\
        MinerU2.0-pipeline & 149 & 16.9\% \\
        MinerU2.0-VLM & 22 & 16.1\% \\
        GPT-4o & 181 & 20.5\% \\
        Qwen2.5-VL-7B-Instruct & 105 & 11.9\% \\
        Experts & 64 & 7.3\% \\
        \bottomrule
    \end{tabular}
\end{table}

Table~\ref{tab:bench_metrics} lists the various performance metrics of \benchopt.

\begin{table}[h]
    \centering
    \caption{Statistics of error type in \benchopt}
    \label{tab:bench_metrics}
    \small
    \begin{tabular}{l c c c}
        \toprule
        \textbf{Element} & \textbf{Metrics} & \textbf{Good Case} & \textbf{BadCase} \\
        \midrule
        Text & Edit & 19.94 & 40.44 \\
        \multirow{2}{*}{Table} & TEDS & 87.23 & 62.84 \\ 
         & STEDS & 97.30 & 67.30 \\
         Equation & Edit & 14.96 & 41.6 \\
        \bottomrule
    \end{tabular}
\end{table}

\noindent\textbf{Error Type Distribution}. In Table~\ref{tab:error_type_bench}, we specifically present the distribution of the different error types in \benchopt.

\begin{table}[!t]
    \centering
    \caption{Statistics of error type in \datasetopt}
    \label{tab:error_type}
    \small
    \begin{tabular}{p{4cm}cc}
        \toprule
        \textbf{Error type} & \textbf{Count} & \textbf{Percentage} \\
        \midrule
        Text misrecognized as formula & 1950 & 0.88\% \\
        Text misrecognized as title & 2950 & 1.33\% \\
        Text misrecognized as table & 1950 & 0.88\% \\
        Superscript citation format in text recognition error & 4440 & 2.00\% \\
        List format recognition error & 4414 & 1.99\% \\
        Title format recognition error & 3343 & 1.51\% \\
        Text paragraph format error & 7525 & 3.39\% \\
        Text redundancy & 11548 & 5.21\% \\
        Text repetition & 2950 & 1.33\% \\
        Extra/missing spaces in text & 10720 & 4.83\% \\
        Text characters lost & 9692 & 4.37\% \\
        Text character recognition error & 9833 & 4.43\% \\
        Text segment lost & 4681 & 2.11\% \\
        Text punctuation recognition error & 8783 & 3.96\% \\
        Inline formula recognition error & 24279 & 10.95\% \\
        Inline formula missed recognition & 6085 & 2.74\% \\
        Inline formula representation style error & 5427 & 2.45\% \\
        Table recognition corruption & 2632 & 1.19\% \\
        Partial table redundancy & 7008 & 3.16\% \\
        Table cell recognition lost & 12552 & 5.66\% \\
        Missing table row/column & 12371 & 5.58\% \\
        Table merged cell error & 11614 & 5.24\% \\
        Table cell content recognition error & 11963 & 5.40\% \\
        Displayed formula misrecognized as text & 3950 & 1.78\% \\
        Displayed formula syntax error & 7707 & 3.48\% \\
        Displayed formula character recognition error & 12065 & 5.44\% \\
        Displayed formula structure recognition error & 11069 & 4.99\% \\
        Partial displayed formula missing & 8216 & 3.71\% \\
        \bottomrule
    \end{tabular}
\end{table}

\begin{table}[h]
    \centering
    \caption{Statistics of error type in \benchopt}
    \label{tab:error_type_bench}
    \small
    \begin{tabular}{p{4cm}cc}
        \toprule
        \textbf{Error type} & \textbf{Count} & \textbf{Percentage} \\
        \midrule
        Text misrecognized as title & 22 & 2.49\% \\
        Text misrecognized as formula & 7 & 0.79\% \\
        Text misrecognized as table & 16 & 1.81\% \\
        Superscript citation format in text recognition error & 11 & 1.25\% \\
        List format recognition error & 34 & 3.85\% \\
        Title format recognition error & 25 & 2.72\% \\
        Text paragraph format error & 31 & 3.51\% \\
        Text redundancy & 39 & 4.42\% \\
        Text repetition & 23 & 2.61\% \\
        Extra/missing spaces in text & 41 & 4.65\% \\
        Text characters lost & 50 & 5.67\% \\
        Text character recognition error & 70 & 7.94\% \\
        Text segment lost & 38 & 4.31\% \\
        Text punctuation recognition error & 32 & 3.63\% \\
        Inline formula recognition error & 32 & 3.63\% \\
        Inline formula missed recognition & 21 & 2.38\% \\
        Inline formula representation style error & 20 & 2.27\% \\
        Table recognition corruption & 26 & 2.95\% \\
        Partial table redundancy & 24 & 2.72\% \\
        Table cell recognition lost & 58 & 6.58\% \\
        Missing table row/column & 50 & 5.67\% \\
        Table merged cell error & 66 & 7.48\% \\
        Table cell content recognition error & 51 & 5.78\% \\
        Displayed formula misrecognized as text & 14 & 1.59\% \\
        Displayed formula syntax error & 39 & 4.42\% \\
        Displayed formula character recognition error & 51 & 5.78\% \\
        Displayed formula structure recognition error & 53 & 6.01\% \\
        Partial displayed formula missing & 52 & 5.90\% \\
        \bottomrule
    \end{tabular}
\end{table}

\clearpage
\section{Template of Chain-of-CheckList(CoCL)}
\label{appendix:cocl_template}
\begin{figure}[h]
  \centering
   \includegraphics[width=1.0\linewidth]{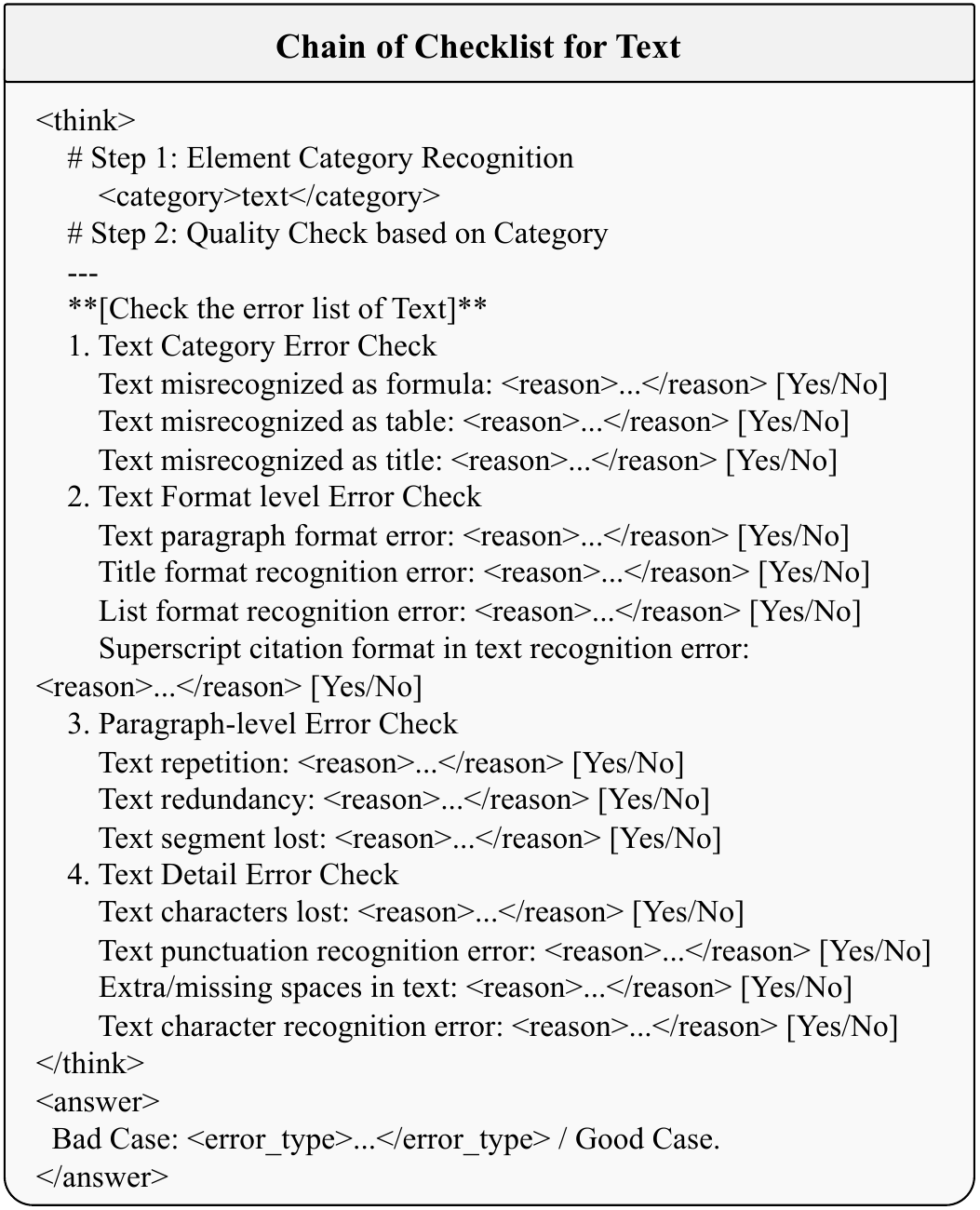}
    \vspace{-6mm}
   \caption{CoCL reasoning template of text.}
   \label{fig:text_cocl}
   \vspace{-6mm}
\end{figure}

\begin{figure}[h]
  \centering
   \includegraphics[width=1.0\linewidth]{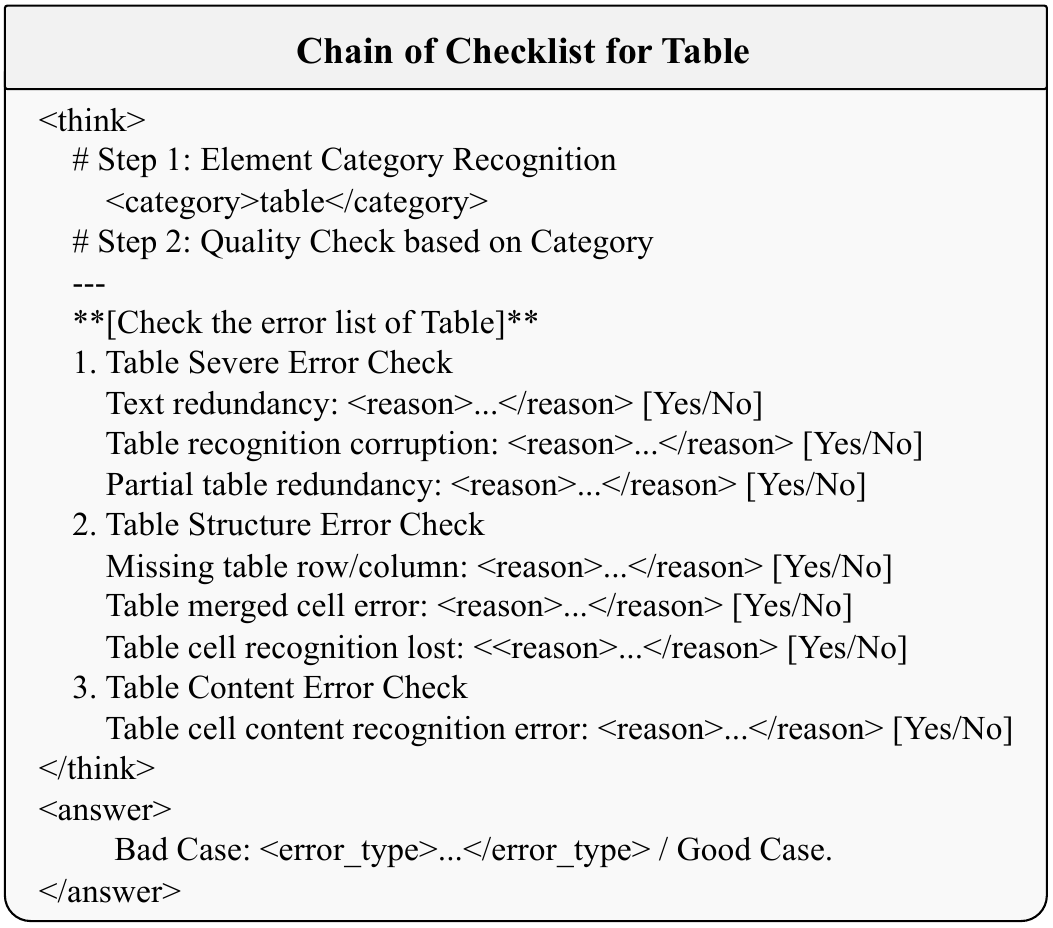}
    \vspace{-6mm}
   \caption{CoCL reasoning template of table.}
   \label{fig:table_cocl}
   \vspace{-6mm}
\end{figure}

\begin{figure}[h]
  \centering
   \includegraphics[width=0.95\linewidth]{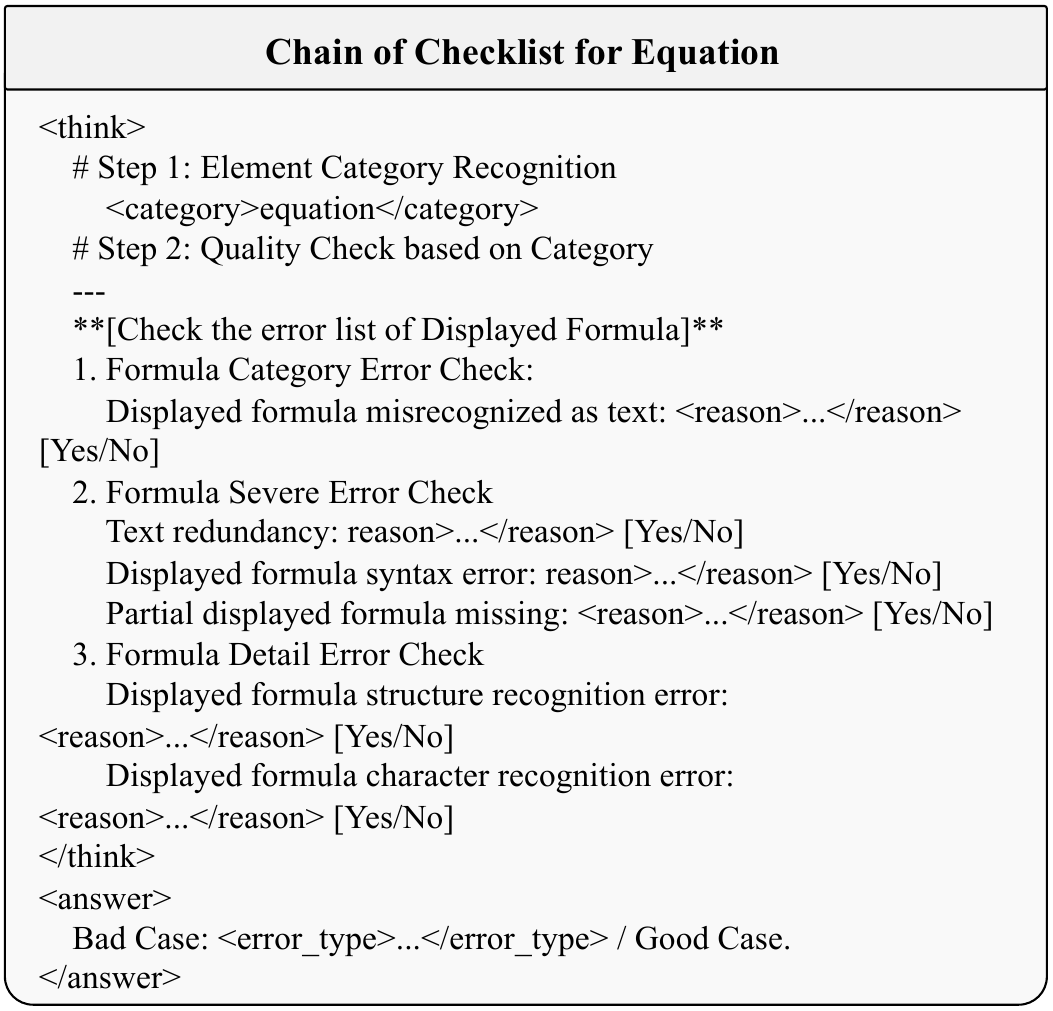}
   \caption{CoCL reasoning template of equation.}
   \label{fig:equation_cocl}
   \vspace{-6mm}
\end{figure}

We illustrate the structured Chain-of-Checklist template in detail in~\cref{fig:text_cocl}, ~\cref{fig:table_cocl}, and~\cref{fig:equation_cocl}.

\clearpage
\section{More Training Details}
\label{appendix:training_details}
The training setups for the two stages are shown in Table~\ref{tab:training_settings}.
We use Qwen2.5-VL-7B-Instruct as the base model and traine \modelopt-7B using 8 $\times$ NVIDIA A100 GPUs.

\modelopt adopts a unified prompt template: \textit{Analyze the quality of OCR results for the given image.$<$ocr\_content$>$...$<$/ocr\_content$>$}. The output format of  is the CoCL format, as shown in Appendix~\ref{appendix:cocl_template}.

In the second stage of training with GRPO, the model's sampling temperature is set to 1.0, with top-k and top-p kept at default, the number of samples $G$ is 8.

\begin{table}[h]
    \centering
    \caption{Training Settings}
    \label{tab:training_settings}
    \begin{tabular}{lcc}
        \toprule
        \textbf{setting} & \textbf{SFT-Stage} & \textbf{GRPO-stage} \\
        \midrule
        \multicolumn{3}{l}{\textit{Dataset}} \\
        \midrule
        \textbf{Data} & \datasetopt & \datasetopt \\
        & 200K & 3K \\
        \textbf{epoch} & 2 & 1 \\
        \midrule
        \multicolumn{3}{l}{\textit{Input of Image}} \\
        \midrule
        \textbf{Max Resolution} & $1280*28*28$ & $1280*28*28$ \\
        \textbf{Token} & $256-1280$ & $256-1280$ \\
        \midrule
        \multicolumn{3}{l}{\textit{Model}} \\
        \midrule
        \textbf{Trainable} & All & LLM \\
        \textbf{Sequence Length} & 16384 & 4096 \\
        \textbf{Torch Dtype} & bf16 & bf16 \\
        \midrule
        \multicolumn{3}{l}{\textit{Train}} \\
        \midrule
        \textbf{Train Type} & full & full \\
        \textbf{Batch Size} & 64 & 64 \\
        \textbf{LR of ViT} & $1.00\text{E-}05$ & - \\
        \textbf{LR of LLM} & $1.00\text{E-}05$ & $1.00\text{E-}06$ \\
        \bottomrule
    \end{tabular}
\end{table}

\clearpage
\section{Prompt of Experiments}
\label{appendix:experiments}
In \cref{tab:performance_prompt}, we present the prompt template used in \cref{tab:performance_prompt}. The prompt employed in refinement experiment is provided in \cref{tab:prompt_refinement}.

\begin{table*}[t]
    \centering
    \small
    \begin{tabular}{p{0.9\linewidth}}
    \toprule
    \textit{Prompt Tamplate} \\
    \midrule
    \textbf{Role} \\
    \quad You are an excellent Quality Assurance Assistant for OCR and Document Parsing results.\\
    \textbf{Task Guidelines} \\
    \quad\quad 1.  I will provide you with an image and the corresponding document parsing result for that image.\\
    \quad\quad 2.  Please carefully inspect the Image and the Parsing Result against the Error Type Definitions provided below, and determine if the Parsing Result contains any errors.\\
    \textbf{Definition of error} \\
    \quad \textcolor{blue}{\{error\_definition\_list\}}\\
    \textbf{Output Format} \\
    \quad \textcolor{blue}{\{output\_format\}} \\
    Start now! The Parsing Result is:\\
    \midrule
    \textit{Detailed description} \\
    \midrule
    \sethlcolor{yellow}\hl{\textbf{error\_definition\_list:}} \\
    Definitions provided in~\cref{fig:app_definition_1},~\cref{fig:app_definition_2} and~\cref{fig:app_definition_3}.\\
    \sethlcolor{yellow}\hl{\textbf{w/ CoT output\_format:}} \\
    \quad 1.  If there are errors (Badcase): \\
    \quad * The output format must strictly be: \\
    \quad\quad $<$think$>$Your thinking and reasoning process$<$/think$>$\\
    \quad\quad $<$answer$>$Badcase.$<$/answer$>$\\
    \quad\quad $<$error\_type$>$error\_type1$<$/error\_type$>$$<$error\_type$>$error\_type2$<$/error\_type$>$ \\
    \quad *The `$<$error\_type$>$` tag must contain all identified error type names.\\
    \quad 2.  If there are no errors (Goodcase):\\
    \quad The output format must strictly be:\\
    \quad\quad $<$think$>$Your thinking and reasoning process$<$/think$>$ \\
    \quad\quad $<$answer$>$Goodcase.$<$/answer$>$\\
    \sethlcolor{yellow}\hl{\textbf{w/o CoT output\_format:}} \\
    \quad 1.  If there are errors (Badcase): \\
    \quad * The output format must strictly be: \\
    \quad\quad $<$answer$>$Badcase.$<$/answer$>$ \\
    \quad\quad $<$error\_type$>$error\_type1$<$/error\_type$>$$<$error\_type$>$error\_type2$<$/error\_type$>$ \\
    \quad *The `$<$error\_type$>$` tag must contain all identified error type names.\\
    \quad Do NOT give any other explanations, notes, or extra text.\\
    \quad 2.  If there are no errors (Goodcase):\\
    \quad *The output format must strictly be:\\
    \quad\quad $<$answer$>$Goodcase.$<$/answer$>$\\
    \quad Do NOT give any other explanations, notes, or extra text.\\
    \bottomrule
    \end{tabular}
    \caption{Prompt used in general-purpose VLMs for document parsing results evaluation.}
    \label{tab:performance_prompt}
\end{table*}   

\begin{table*}[t]
    \centering
    \small
    \begin{tabular}{p{0.9\linewidth}}
    \toprule
    \textbf{Role} \\
    \quad You are an expert in document image analysis and correction with a highly developed visual understanding ability, specialized in precise structural analysis of complex document elements, including tables and mathematical formulas. Your task is to perform precise and complete corrections on the original OCR results based on the provided document image, the type of elements in the image, the original OCR output, \textcolor{blue}{and the quality control feedback}, ultimately delivering the highest quality parsed content and accurate element structure. \\
    \textbf{Task Guidelines} \\
    \quad  - Input Items \\
    \quad \quad - Document Image (Visual Input): A cropped image of a document of [category] type.\\
    \quad \quad - Original OCR Result: The raw output from an OCR model on the document image.\\
    \quad \quad \textcolor{blue}{- Quality Control Feedback: Errors and their explanations provided by an error-correction model.}\\
    \quad \textcolor{blue}{- Correction Principles}\\
    \quad \quad  \textcolor{blue}{- Prioritize Quality Control Feedback: Corrections must be made based on the errors and reasons indicated in the quality control feedback, ensuring that the corrected content aligns with the suggestions and guidance from the feedback.}\\
    \quad \quad \textcolor{blue}{- Visual and Structural Verification:Even with feedback, you must carefully examine the document image to ensure that all corrections are completely faithful to the image, ensuring no new errors are introduced and that no potentially correct content is modified.}\\
    \quad - Correction Process\\
    \quad \quad - Analyze the document image to identify text, tables, mathematical formulas, or other elements.\\
    \quad \quad - Review the original OCR result and evaluate potential recognition errors or quality issues in the OCR model output. If the OCR result (pred) is empty, directly analyze the content in the image and perform recognition.\\
    \quad \quad \textcolor{red}{- Determine if there are errors in the content. If errors exist, perform corrections to obtain a more accurate output result, ensuring no new errors are introduced; otherwise, directly output the original result.}\\
    \quad \quad \textcolor{blue}{- Based on the quality control feedback, check for errors in the OCR result and make corrections. The corrections should consider language, context, and visual features.}\\
    \quad \quad \textcolor{blue}{- erform visual and structural verification to ensure all corrected content and structure matches the original document image without introducing new errors.}\\
    \quad \quad - Based on the element type (text, table, or mathematical formula), output the corrected content in the specified format.\\
    \textbf{Output Format} \\
    \quad - The model must wrap its internal thought process using the `$<$think$>$` tag and the final corrected result using the `$<$answer$>$` tag.\\
    \quad Output Format within `$<$answer$>$` Tag:\\
    \quad\quad - Text: If the document element is a text block, output the corrected plain text, ensuring grammar, spelling, formatting, etc., align with the overall style of the document.\\
    \quad\quad -*Tables (Strict HTML Format): If the document element is a table, output the corrected table in Standard HTML format (`$<$table$>$`, `$<$tr$>$`, `$<$td$>$`, `$<$th$>$`). You must ensure the following structural elements are accurately reflected based on the document image:\\
    \quad\quad 1. Cell Merging Correctly use `colspan` and `rowspan` attributes to reproduce merged cells exactly as seen in the image.\\
    \quad\quad 2. Table Headers Use `$<$th$>$` tags for all header cells (both column and row headers) to correctly distinguish them from data cells (`$<$td$>$`).\\
    \quad\quad 3. No Styling: Strictly avoid generating any CSS styles or attributes related to visual presentation (e.g., `width`, `height`, `align`, `style`). Focus only on the logical data structure.\\
    \quad\quad - Mathematical Formulas: If the document element is a mathematical formula, output the corrected formula in LaTeS format. Pay attention to symbols, operators, and the accuracy of the formula structure. \\
    \\
    Let's begin!  \\
    \bottomrule
    \end{tabular}
    \caption{The prompt template supports three refinement settings, with green highlighting instructions specific to the unguided(w/o G) and binary-guided(w/ BG) modes, and blue indicating content exclusive to the detailed quality assessment guidance(w/ DG) setting.}
    \label{tab:prompt_refinement}
\end{table*}   

\clearpage
\section{Case Study}
\label{appendix:case}
\begin{figure*}[h]
  \centering
   \includegraphics[width=1.0\linewidth]{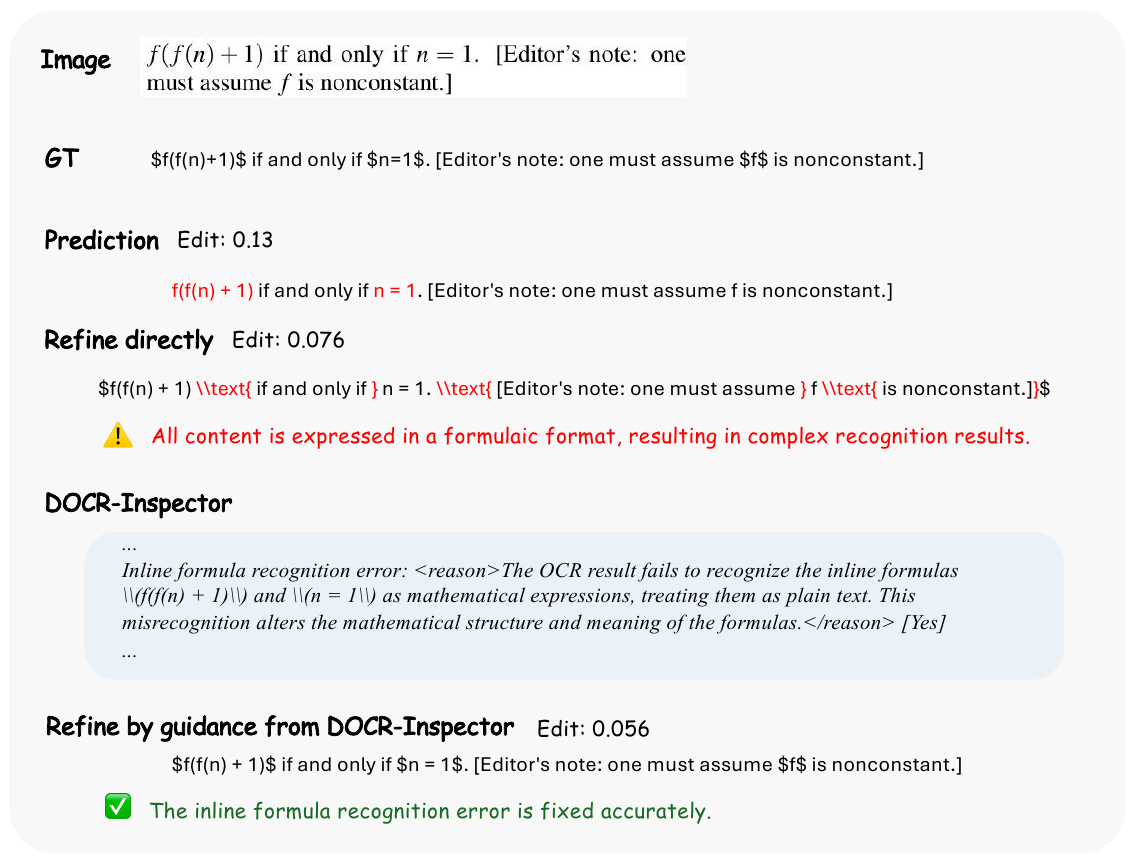}
    \vspace{-6mm}
   \caption{A text case showing the initial prediction, the quality feedback from \modelopt-7B, and outputs from two refinement strategies.}
   \label{fig:app_case_1}
   \vspace{-6mm}
\end{figure*}

\begin{figure*}[h]
  \centering
   \includegraphics[width=1.0\linewidth]{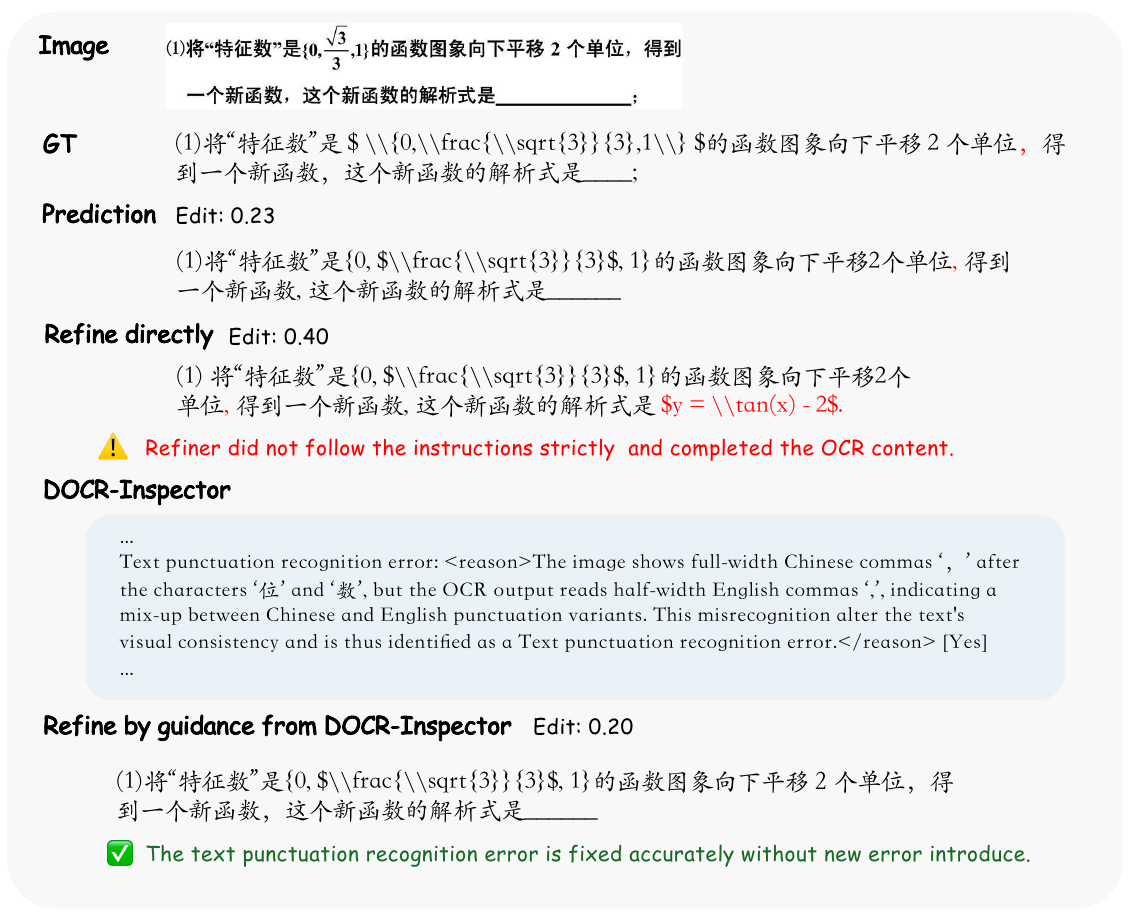}
    \vspace{-6mm}
   \caption{A text case showing the initial prediction, the quality feedback from \modelopt-7B, and outputs from two refinement strategies.}
   \label{fig:app_case_2}
   \vspace{-6mm}
\end{figure*}

\begin{figure*}[h]
  \centering
   \includegraphics[width=1.0\linewidth]{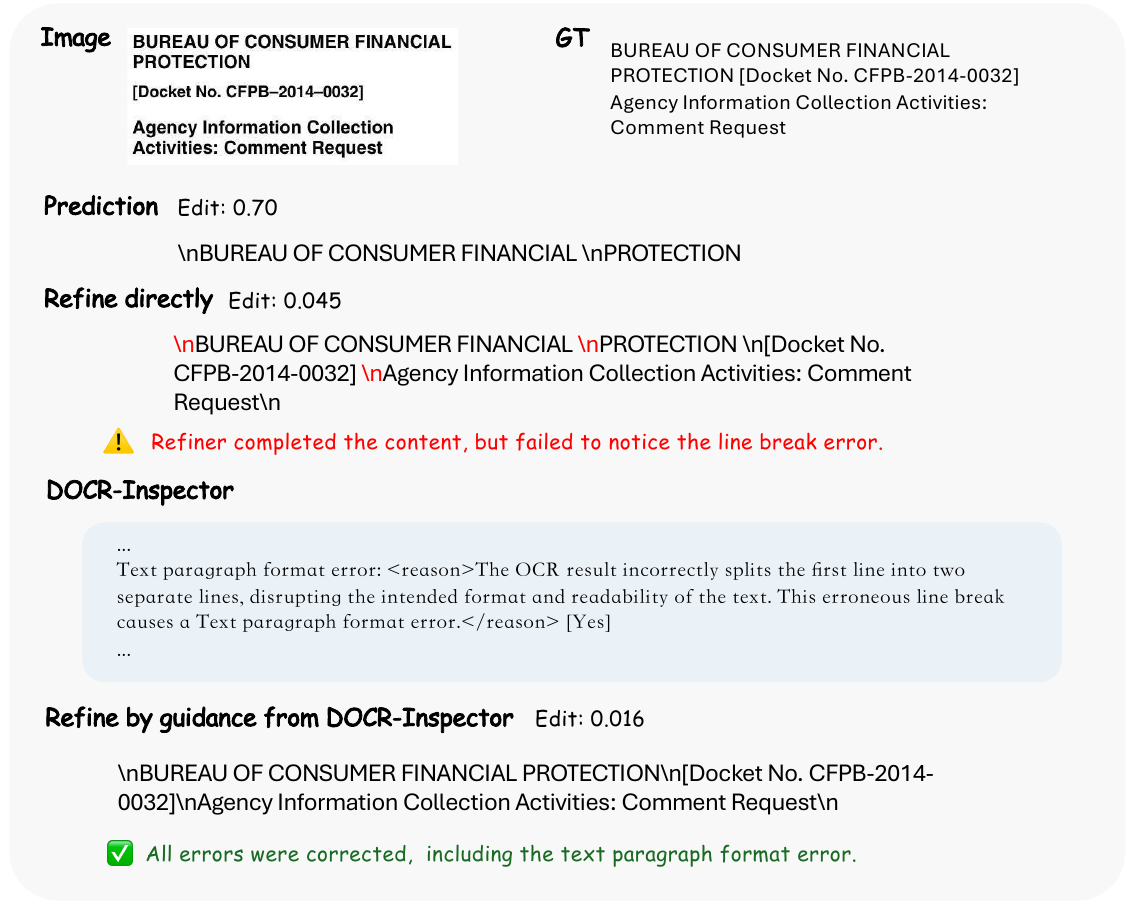}
    \vspace{-6mm}
   \caption{A text case showing the initial prediction, the quality feedback from \modelopt-7B, and outputs from two refinement strategies.}
   \label{fig:app_case_3}
   \vspace{-6mm}
\end{figure*}

\begin{figure*}[h]
  \centering
   \includegraphics[width=1.0\linewidth]{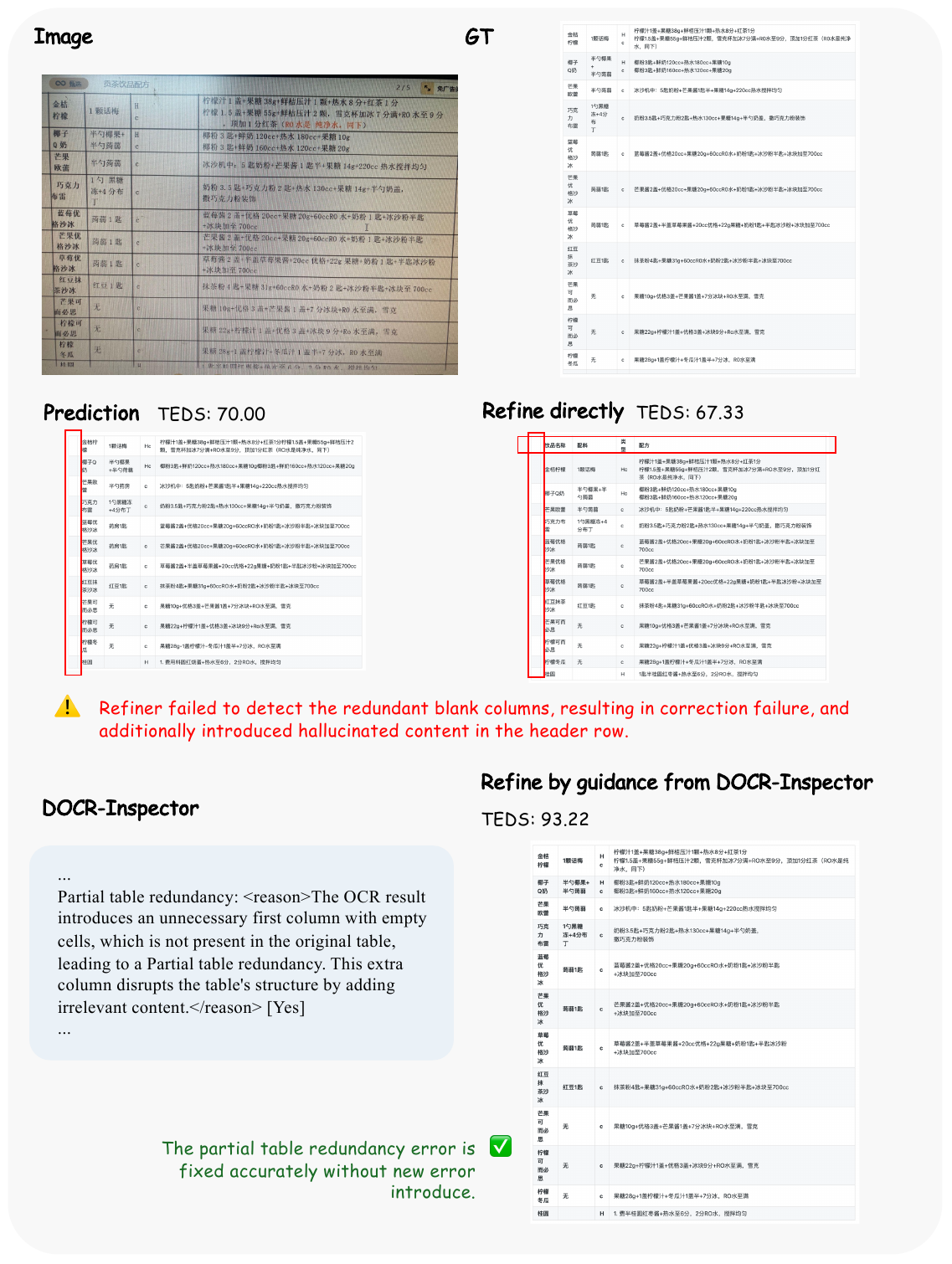}
    \vspace{-6mm}
   \caption{A table case showing the initial prediction, the quality feedback from \modelopt-7B, and outputs from two refinement strategies.}
   \label{fig:app_case_4}
   \vspace{-6mm}
\end{figure*}

\begin{figure*}[h]
  \centering
   \includegraphics[width=1.0\linewidth]{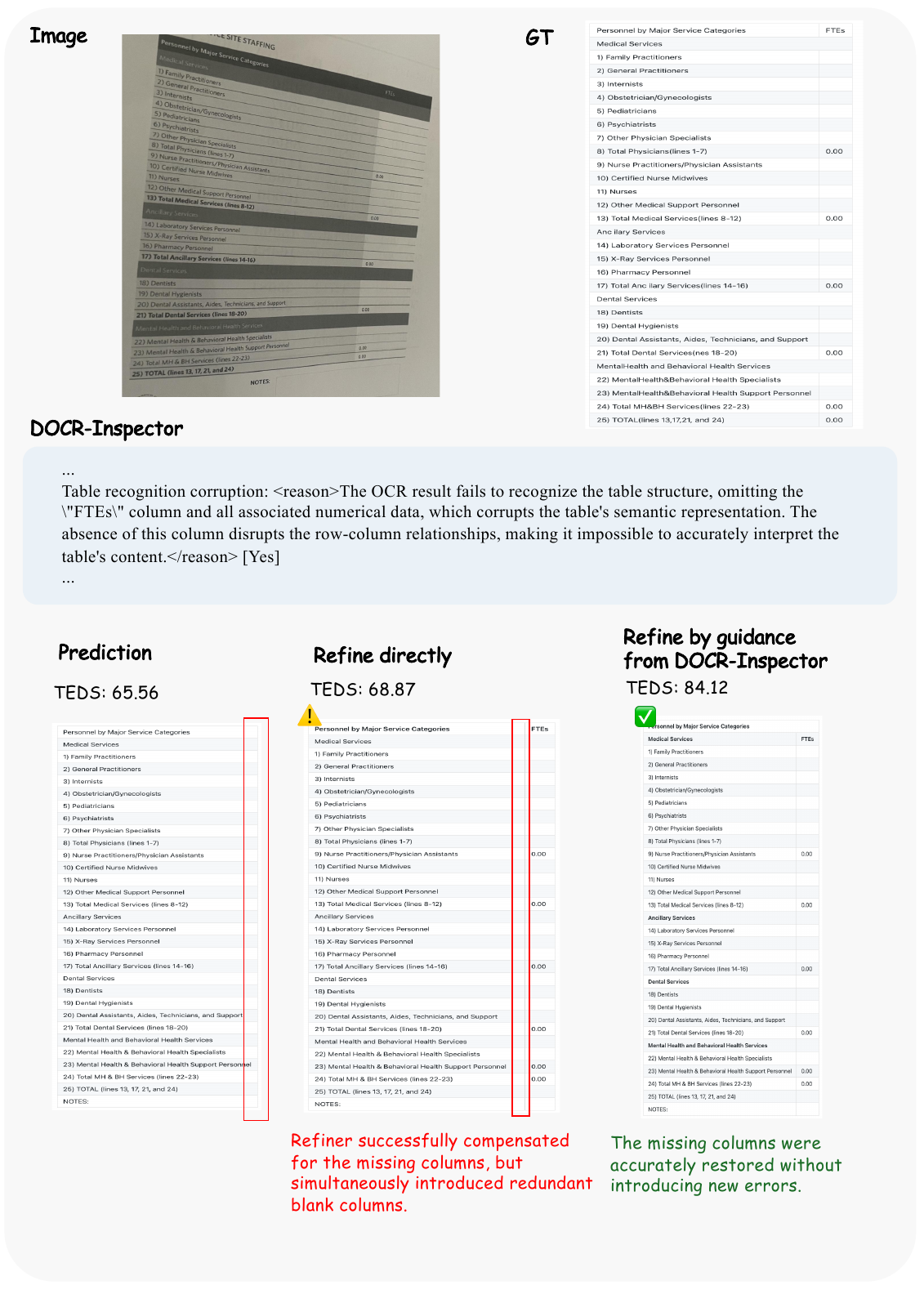}
    \vspace{-6mm}
   \caption{A table case showing the initial prediction, the quality feedback from \modelopt-7B, and outputs from two refinement strategies.}
   \label{fig:app_case_5}
   \vspace{-6mm}
\end{figure*}

\begin{figure*}[h]
  \centering
   \includegraphics[width=1.0\linewidth]{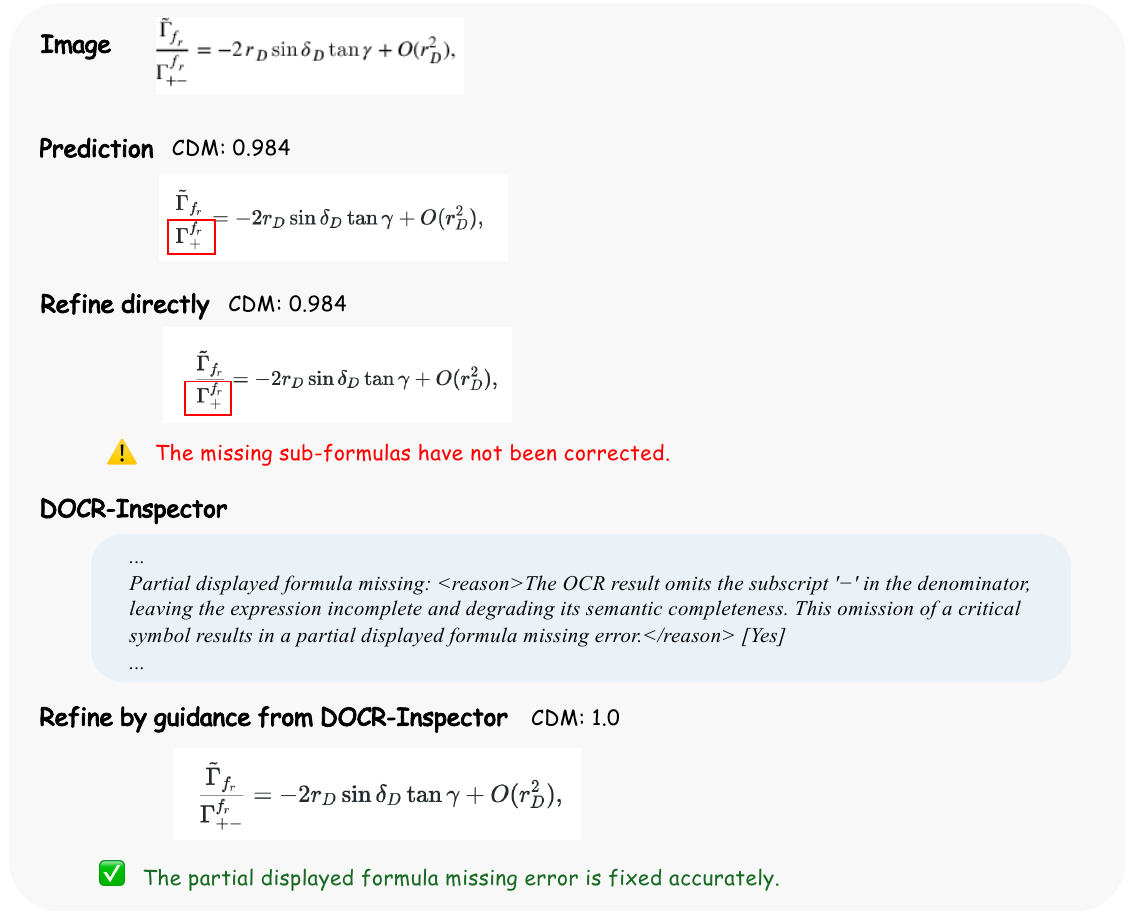}
    \vspace{-6mm}
   \caption{A equation case showing the initial prediction, the quality feedback from \modelopt-7B, and outputs from two refinement strategies.}
   \label{fig:app_case_6}
   \vspace{-6mm}
\end{figure*}

\begin{figure*}[h]
  \centering
   \includegraphics[width=1.0\linewidth]{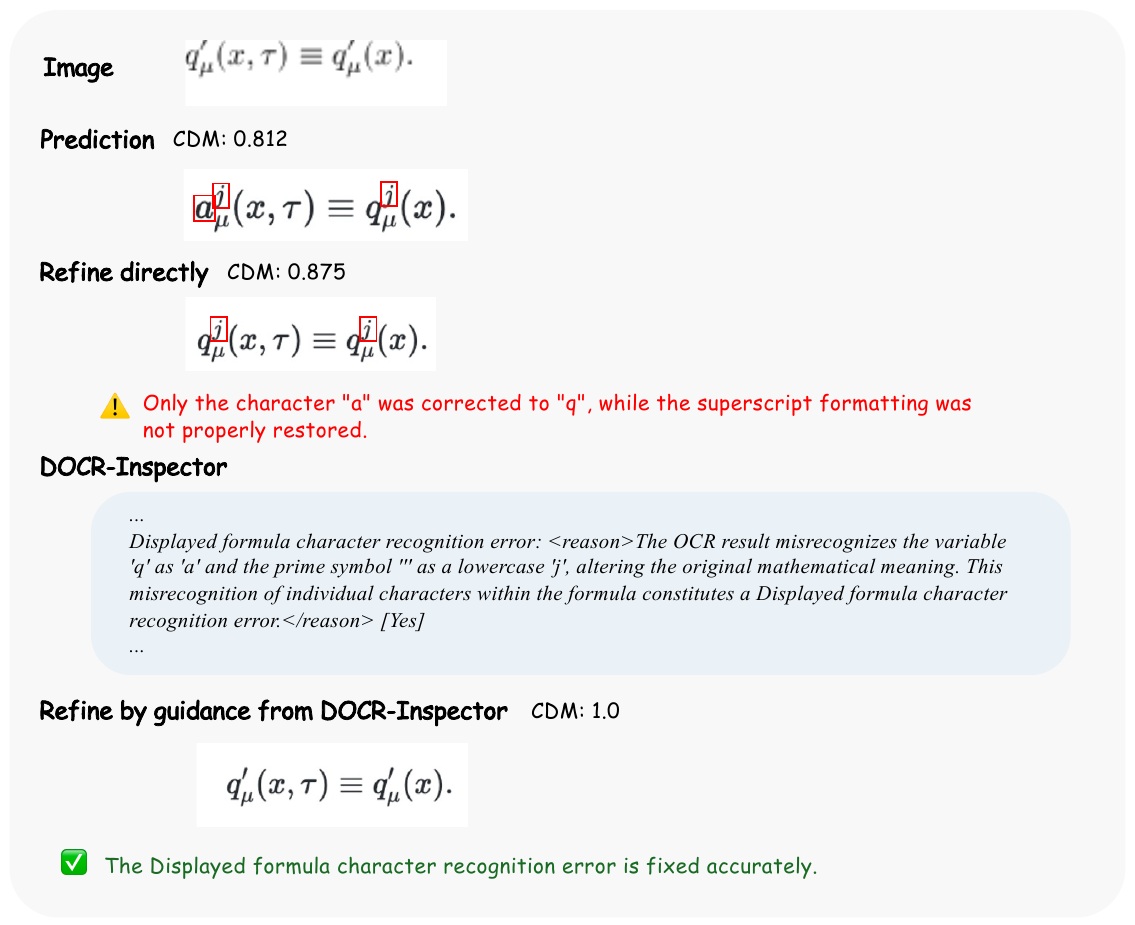}
    \vspace{-6mm}
   \caption{A equation case showing the initial prediction, the quality feedback from \modelopt-7B, and outputs from two refinement strategies.}
   \label{fig:app_case_7}
   \vspace{-6mm}
\end{figure*}

Figures~\ref{fig:app_case_1} to~\ref{fig:app_case_7} show selected cases from the refinement experiments. For each case, we present the image, the ground truth, and the prediction from either dots.ocr or MonkeyOCR-3B-pro. We also provide the direct refinement result, the quality feedback from \modelopt-7B, and the refinement result guided by this feedback. We employ Qwen2.5-VL-72B-Instruct as the refiner.

\end{document}